%% file: main.tex
\documentclass[11pt, a4paper, logo, copyright, nonumbering]{Bohr_Sci}

\usepackage{graphicx}%
\usepackage{multirow}%
\usepackage{amsmath,amssymb,amsfonts}%
\usepackage{amsthm}%
\usepackage{mathrsfs}%
\usepackage[title]{appendix}%
\usepackage{xcolor}%
\usepackage{textcomp}%
\usepackage{manyfoot}%
\usepackage{booktabs}%
\usepackage{algorithm}%
\usepackage{algorithmicx}%
\usepackage{algpseudocode}%
\usepackage{listings}%
\usepackage{mathtools}
\usepackage{tabularx}
\usepackage[most]{tcolorbox}
\usepackage{letltxmacro}
\usepackage{xstring}
\usepackage{environ}
\usepackage{array}

\usepackage[most]{tcolorbox}
\usepackage{enumitem}

\definecolor{agentbg}{RGB}{245,248,252}      
\definecolor{agentborder}{RGB}{210,220,235}  
\definecolor{agenttitle}{RGB}{40,60,110}     

\newtcolorbox{agentcard}[1][]{
  enhanced,
  breakable,
  colback=agentbg,
  colframe=agentborder,
  coltitle=agenttitle,
  boxrule=0.6pt,
  arc=3pt,
  left=6pt,right=6pt,top=6pt,bottom=6pt,
  title={#1},
  fonttitle=\bfseries\large,
}

\usepackage{tikz}
\usetikzlibrary{arrows.meta, positioning, fit, shapes.geometric, shapes.misc}

\newcommand{\agentfield}[2]{\textbf{#1:}~#2\\}

\theoremstyle{thmstyleone}%
\theoremstyle{thmstyletwo}%

\theoremstyle{thmstylethree}%

\usepackage[capitalize,noabbrev]{cleveref}

\definecolor{wikiblue}{HTML}{1195DB}
\definecolor{Grokipediaorange}{HTML}{F4A324} 
\definecolor{sciencepediagreen}{HTML}{27B57A}
\definecolor{baselinecolor}{HTML}{DC5151}

\newtcolorbox{wikipediaarticle}[1][Wikipedia Article]{
    colback=wikiblue!10!white,
    colframe=wikiblue,
    fonttitle=\bfseries,
    breakable,
    title={#1}
}

\newtcolorbox{sciencepediaarticle}[1][Sciencepedia Article]{
    colback=sciencepediagreen!10!white,
    colframe=sciencepediagreen,
    fonttitle=\bfseries,
    breakable,
    title={#1}
}

\newtcolorbox{llmbaseline}[1][Baseline LLM Article]{
    colback=baselinecolor!10!white,
    colframe=baselinecolor,
    fonttitle=\bfseries,
    breakable,
    title={#1}
}

\newtcolorbox{grokipediaarticle}[1][Grokipedia Article]{
    colback=Grokipediaorange!10!white,
    colframe=Grokipediaorange,
    fonttitle=\bfseries,
    breakable,
    title={#1}
}

\title{Bohrium + SciMaster: Building the  Infrastructure and Ecosystem for Agentic Science at Scale}

\author{
Linfeng Zhang\textsuperscript{1,2},
Siheng Chen\textsuperscript{3},
Yuzhu Cai\textsuperscript{4},
Jingyi Chai\textsuperscript{3},
Junhan Chang\textsuperscript{1,5},
Kun Chen\textsuperscript{6},
Zhi X. Chen\textsuperscript{5},
Zhaohan Ding\textsuperscript{1},
Yuwen Du\textsuperscript{3},
Yuanpeng Gao\textsuperscript{1,5},
Yuan Gao\textsuperscript{1},
Jing Gao\textsuperscript{1,3,9},
Zhifeng Gao\textsuperscript{1},
Qiangqiang Gu\textsuperscript{10},
Yanhui Hong\textsuperscript{1},
Yuan Huang\textsuperscript{1},
Xi Fang\textsuperscript{1},
Xiaohong Ji\textsuperscript{1},
Guolin Ke\textsuperscript{1},
Zixing Lei\textsuperscript{3},
Xinyu Li\textsuperscript{2},
Yongge Li\textsuperscript{1},
Ruoxue Liao\textsuperscript{1},
Hang Lin\textsuperscript{11},
Xiaolu Lin\textsuperscript{1},
Yuxiang Liu\textsuperscript{2},
Xinzijian Liu\textsuperscript{1},
Zexi Liu\textsuperscript{3},
Jintan Lu\textsuperscript{1},
Tingjia Miao\textsuperscript{3},
Haohui Que\textsuperscript{1,7,8},
Weijie Sun\textsuperscript{1},
Yanfeng Wang\textsuperscript{3},
Bingyang Wu\textsuperscript{1,2},
Tianju Xue\textsuperscript{12},
Rui Ye\textsuperscript{3},
Jinzhe Zeng\textsuperscript{10},
Duo Zhang\textsuperscript{5,2,1},
Jiahui Zhang\textsuperscript{1},
Linfeng Zhang\textsuperscript{3},
Tianhan Zhang\textsuperscript{4},
Wenchang Zhang\textsuperscript{12},
Yuzhi Zhang\textsuperscript{1},
Zezhong Zhang\textsuperscript{1,2},
Hang Zheng\textsuperscript{1},
Hui Zhou\textsuperscript{1},
Tong Zhu\textsuperscript{7,8},
Xinyu Zhu\textsuperscript{3},
Qingguo Zhou\textsuperscript{1,2},
Weinan E\textsuperscript{2,5}
\\
\textsuperscript{1} DP Technology, Beijing, China \\
\textsuperscript{2} AI for Science Institute, Beijing, China \\
\textsuperscript{3} Shanghai Jiao Tong University, Shanghai, China \\
\textsuperscript{4} Beihang University, Beijing, China \\
\textsuperscript{5} Peking University, Beijing, China \\
\textsuperscript{6} Institute of Theoretical Physics, Chinese Academy of Sciences, Beijing, China \\
\textsuperscript{7} Shanghai Innovation Institute, Shanghai, China \\
\textsuperscript{8} East China Normal University, Shanghai, China \\
\textsuperscript{9} Zhongguancun Academy, Beijing, China \\
\textsuperscript{10} University of Science and Technology of China, Hefei, China \\
\textsuperscript{11} Tongji University, Shanghai, China \\
\textsuperscript{12} The Hong Kong University of Science and Technology, Hong Kong, China
}

\begin{document}

\maketitle
\thispagestyle{firststyle}

\definecolor{Scimaster_Blue}{HTML}{004098}
\centerline{\color{Scimaster_Blue}\fontsize{15pt}{14pt}\selectfont\textbf{Abstract}}
\vspace{2ex}

\absfont
\noindent{AI agents are emerging as a practical way to run multi-step scientific workflows that interleave reasoning with tool use and verification, pointing to a shift from isolated AI-assisted steps toward \emph{agentic science at scale}. This shift is increasingly \emph{feasible}, as scientific tools and models can be invoked through stable interfaces and verified with recorded execution traces, and increasingly \emph{necessary}, as AI accelerates scientific output and stresses the peer-review and publication pipeline, raising the bar for traceability and credible evaluation.}

\noindent{However, scaling agentic science remains difficult: workflows are hard to observe and reproduce; many tools and laboratory systems are not agent-ready; execution is hard to trace and govern; and prototype ``AI Scientist'' systems are often bespoke, limiting reuse and systematic improvement from real workflow signals.}

\noindent{We argue that scaling agentic science requires an \emph{infrastructure-and-ecosystem} approach, instantiated in \emph{Bohrium+SciMaster}. Bohrium acts as a managed, traceable hub for AI4S assets---akin to a ``HuggingFace of AI for Science''---that turns diverse scientific data, software, compute, and laboratory systems into agent-ready capabilities. SciMaster orchestrates these capabilities into long-horizon scientific workflows, on which scientific agents can be composed and executed. Between infrastructure and orchestration, a \emph{scientific intelligence substrate} organizes reusable models, knowledge, and components into executable building blocks for workflow reasoning and action, enabling composition, auditability, and improvement through use.}

\noindent{We demonstrate this stack with eleven representative master agents in real workflows, achieving orders-of-magnitude reductions in end-to-end scientific cycle time and generating execution-grounded signals from real workloads at multi-million scale. Together, this work points toward \emph{Science-as-a-Service}: scientific workflows become progressively more executable, observable, and continuously improvable, while remaining anchored in traceable validation and human judgment.}

\noindent{\textbf{Keywords}: agentic science at scale; scientific production; agent-ready infrastructure; scientific workflows; Science-as-a-Service}

\tableofcontents

\section{Introduction}
\label{sec:intro}
\input{sec-1-intro}

\section{Bohrium: Agent-Ready Infrastructure for Scientific Production}
\label{sec:bohrium}
\input{sec-2-bohrium}

\section{Scientific Intelligence Substrate: Models, Knowledge, and Communities}
\label{sec:intelligence_substrate}
\input{sec-3-intelligence_substrate}

\section{SciMaster and a Federated Ecosystem of Scientific AI Agents}
\label{sec:scimaster}
\input{sec-4-scimaster}

\section{Master Agents in Practice: Eleven Case Studies}
\label{sec:master-agents}
\input{sec-5-casestudy}

\section{The Emergence of a Community-Scale Scientific Flywheel}
\label{sec:flywheel}
\input{sec-6-flywheel}

\section{Conclusion and Outlook}
\label{sec:outlook}
\input{sec-7-conclusion}

\newpage

\section*{Acknowledgements}

We thank the engineering and product teams who built and operated the Bohrium+SciMaster infrastructure, enabling reliable large-scale scientific execution across reading, computing, and experimentation. We are also grateful to the broader community and ecosystem partners who contributed tools, models, workflows, and feedback, and whose sustained participation made this work possible.

\bibliography{main}
\bibliographystyle{bst/sn-nature}


\newpage

\begin{appendices}
\section{Master agent cards}
\label{app:agent-cards}
\input{sec-append}
\end{appendices}

\end{document}

%% file: sec-1-intro.tex
In recent years, AI-driven scientific discovery has moved beyond isolated, task-specific applications toward a broader opportunity: treating parts of scientific work as \emph{executable} and \emph{improvable} production. While early \emph{AI for Science} (AI4S) advances delivered striking progress on individual capabilities---protein structure prediction~\cite{jumper2021alphafold2}, materials screening~\cite{merchant2023gnome}, and simulation acceleration~\cite{zhang2018deepmd}---scientific progress is usually realized through end-to-end workflows that connect modelling, execution, and validation.

A new class of agentic systems suggests that such workflows can increasingly be \emph{run} as multi-step cycles that interleave reasoning with tool use and verification, while producing inspectable evidence trails. For example, Kosmos runs long-horizon literature search and analysis to generate traceable, citation-grounded scientific reports~\cite{mitchener2025kosmos}; Robin connects hypothesis generation with lab-in-the-loop experiment planning and execution in engineered settings~\cite{ghareeb2025robin}; and the AI co-scientist from Google/DeepMind uses structured multi-agent generate--debate--evolve procedures to propose biomedical hypotheses, some experimentally validated~\cite{natarajan2025aicoscientist}. These systems collectively indicate that scientific agency is no longer purely speculative.

Herein, we study agentic science \emph{at scale}: scientific production in which \emph{tools, corpora, and workflows} are exposed as \emph{agent-ready, traceable capabilities}, and in which \emph{many agents and many human researchers} can concurrently \emph{execute, inspect, validate, and reuse} end-to-end workflows spanning Reading, Computing, and Experiment on a \emph{shared substrate}. \emph{At scale} thus refers not only to longer-horizon autonomy, but also to \emph{platform-scale participation and reuse}: shared interfaces and execution traces make methods, intermediate artifacts, and validation outcomes portable across tasks and contributors, enabling systematic comparison and iterative improvement rather than one-off demonstrations. The remaining question is how to move from carefully engineered agent prototypes to \emph{shared environments} that can sustain reliable execution, evaluation, and reuse across diverse workflows, backends, and participants.

\subsection{Why now: Why scaling agentic science is necessary and feasible}
\label{subsec:why-now}

The opportunity above is not only attractive; it is increasingly \emph{necessary}. As AI accelerates the generation of hypotheses, analyses, and manuscripts, the peer-review and publication pipeline faces increasing stress, making traceability, reproducibility, and workflow-grounded evaluation first-order requirements for sustaining trust at community scale. More broadly, modern scientific work is increasingly collaborative and pipeline-based: rapidly expanding literature and data, rising experimental costs, and complex multi-tool workflows make ``small-workshop'' practices brittle, hard to reproduce, and difficult to scale across teams and institutions, and leave workflows hard to trace, replay, and evaluate across settings.

At the same time, scaling agentic science is increasingly \emph{feasible} because several system conditions are maturing. First, more scientific tools and models can be invoked through stable interfaces, scheduled on shared compute, and verified with recorded execution traces, turning individual capabilities into reusable workflow components. Second, tool ecosystems are being standardized and evaluated around agent use (e.g., ChemAgent/ChemToolBench~\cite{wu2025chemagent,ramos2024llmchemreview}, X-Master~\cite{chai2025scimaster}), making interface contracts, provenance, and validation gates first-class concerns. Third, experimentation is becoming more programmable through SDLs and laboratory operating systems (e.g., ChemOS~2.0~\cite{sim2024chemos2}, UniLabOS~\cite{unilabos2025internal}), enabling ``doing and verifying'' to be orchestrated and inspected within larger workflows.

Together, these trends create an inflection point: end-to-end scientific cycles can now be run with richer automation and evidence trails, making shared, agent-ready environments a first-order requirement rather than an engineering detail.

\subsection{Four bottlenecks to scaling agentic science}
\label{subsec:bottlenecks}

When agents are embedded into real scientific environments, they still frequently fail to act reliably end-to-end. We argue the main bottleneck is no longer model capability alone, but the absence of shared, explicitly agent-ready environments for scientific production. We summarize this obstacle as a \emph{tooling gap}: scientific actions are not consistently exposed as dependable workflow building blocks; execution is not uniformly traceable and governable; and long-horizon workflows are difficult to run, inspect, and improve across many tasks and participants. Concretely, we observe four intertwined bottlenecks.

\textbf{(1) Weak workflow observability and reproducibility.} Intermediate states are often implicit, making it difficult to inspect what happened, compare alternative runs, or replay a workflow under controlled conditions. This leads to ``works-once'' pipelines and fragile integrations that do not accumulate into shared, reusable assets.

\textbf{(2) Tools are not agent-ready, so plans do not reliably become actions.} Scientific software and laboratory systems are typically designed for humans: non-standard interfaces, hidden state, and fragile I/O conventions prevent reliable autonomous execution. By \emph{agent-ready} we mean capabilities that expose stable programmatic interfaces, support safe composition under explicit contracts, and produce traces that can be inspected, validated, and replayed.

\textbf{(3) From prototypes to production: portability, comparability, and reuse break down.} Scaling exposes workflow-level gaps: portability across institutions and backends; comparability via standardized traces and metrics for regression and system evaluation; and reuse via packaging and explicit contracts that make capabilities composable building blocks.

\textbf{(4) Weak flywheels: execution rarely drives continual improvement.} Without execution-grounded feedback loops, improvements remain local. A scalable system must capture workflow-level signals (success/failure, time, cost, constraint satisfaction, validation evidence) on a shared substrate so that iteration can propagate across interfaces, orchestration policies, knowledge, and models.

Rather than ranking systems by nominal autonomy, we consider a continuum from AI co-pilots and AI co-scientists to more autonomous AI Scientists under human-specified goals and values. Our emphasis is on the shared environments required for any point along this continuum to operate reliably, and on measurable impact on real workflows with broad human participation, measured by workflow-level metrics such as cycle time, robustness, and validation outcomes.

\subsection{Our approach: Bohrium+SciMaster, Infrastructure+Ecosystem}
\label{subsec:approach}

To close the tooling gap, we propose Bohrium+SciMaster as an \emph{infrastructure-and-ecosystem} approach to building a \emph{shared, agent-ready environment} for scientific production: (i) an agent-ready execution substrate that makes Reading, Computing, and Experiment executable as services with traceable and governed execution; (ii) a reusable scientific intelligence substrate spanning models, knowledge, and community-contributed capabilities; and (iii) orchestration and feedback loops that enable long-horizon workflows to be executed, validated, compared, and iteratively refined at ecosystem scale.

Bohrium provides the execution substrate for scientific production, integrating \emph{Science Navigator} for traceable access to literature- and patent-scale evidence, \emph{Lebesgue} for unified compute orchestration across cloud/HPC/AI accelerators, and \emph{UniLabOS} for programmable laboratory execution~\cite{unilabos2025internal}. On this substrate, tools, workflows, datasets, and models become reliably callable, schedulable, observable, and accountable by agents and humans alike, enabling comparable execution across heterogeneous backends. A key platform element is a capability and tool registry that supports standardized packaging and governed execution of large collections of reusable tools and workflows.

On top of Bohrium and the intelligence substrate, \emph{SciMaster} serves as an orchestration runtime for agentic scientific production. It plans and coordinates multi-step, multi-agent work; schedules tools and resources across reading, computing, and experiment services; and maintains long-term state and memory for scientific problem entities. Around SciMaster, domain-specialized agents, tools, and human collaborators form federated ecosystems of scientific AI agents that can be composed, audited, and progressively improved.

Between infrastructure and orchestration sits a scientific intelligence substrate. It includes a hierarchy of scientific models---from general-purpose scientific foundation models (e.g., \emph{Innovator}) to domain foundation models and pipeline/application models---and a knowledge substrate that supports reuse and composability in scientific reasoning. Within our ecosystem, \emph{SciencePedia} provides an open, collaborative scientific knowledge base and associated reasoning components~\cite{sciencepedia_lcot}. In parallel, open AI4S communities such as \emph{DeepModeling} contribute widely used open-source assets---including models, solvers, tools, workflows, evaluation suites, and agent components---that can be integrated into agent-ready workflows on top of the same execution substrate.

Beyond exposing technical capabilities, this stack enables community-scale feedback loops grounded in observable usage and execution traces. In the current deployment, Bohrium supports usage-based charging and subscription plans for certain services and capabilities, with costs and outcomes recorded alongside execution traces. Beyond these mechanisms, we are exploring additional incentive designs that may better align usage signals with value creation and rewards for community-developed tools, models, and laboratory capabilities (Section~\ref{sec:flywheel}). Because usage, costs, outcomes, and validation traces are observed on a shared substrate, improvements can propagate backward across interfaces, orchestration policies, knowledge bases, and models. We view this direction as a practical path toward \emph{Science-as-a-Service}: a mode of scientific production in which execution, evaluation, and improvement are continuously supported by shared infrastructure and an evolving ecosystem.

Figure~\ref{fig:overview} summarizes the stack around Bohrium+SciMaster.

\begin{figure*}[!t]
\centering
\includegraphics[width=\textwidth]{figures/bohrium_scimaster_overview.png}
\caption{\textbf{Infrastructure and ecosystem around Bohrium+SciMaster.} Bohrium turns bare scientific assets (data, software, compute, and laboratory equipment) into agent-ready capabilities for Reading, Computing, and Experiment with unified interfaces, observability, and governance. It also hosts a capability and tool registry for standardized packaging and governed execution of large collections of reusable tools and workflows. Above this execution substrate, a scientific intelligence substrate provides reusable building blocks for scientific reasoning and action, including a hierarchy of scientific models (e.g., Innovator, domain foundation models, and pipeline/application models) and a knowledge substrate (SciencePedia). Open AI4S communities---represented here by DeepModeling---contribute across the stack and provide reusable open-source assets that can be integrated into agent-ready workflows on top of shared infrastructure. SciMaster orchestrates these capabilities into long-horizon, tool-augmented, multi-agent workflows, while execution traces and distributed validation signals support continual refinement at ecosystem scale.}
\label{fig:overview}
\end{figure*}

\subsection{Contributions and organization}
\label{subsec:contrib}

This paper makes three contributions: (1) a production-centric, system-level framing of \emph{agentic science at scale} that emphasizes reliable end-to-end execution, comparable evaluation, and broad human participation rather than isolated model autonomy; (2) an infrastructure-and-ecosystem stack centered on Bohrium that turns scientific assets into agent-ready production capabilities with traceable and governed execution, supporting platform-scale reuse and community participation; and (3) SciMaster as an orchestrator for long-horizon agentic scientific production, together with eleven representative master agents that already exhibit substantial reductions in several end-to-end scientific workflows.

The remainder of the paper is organized as follows. Section~\ref{sec:bohrium} introduces Bohrium as the execution substrate for production-scale agentic science. Section~\ref{sec:intelligence_substrate} describes the scientific intelligence substrate---a hierarchy of models, knowledge bases, and communities. Section~\ref{sec:scimaster} describes SciMaster and its federated ecosystem of scientific AI agents. Section~\ref{sec:master-agents} presents eleven representative master agents built on this stack. Section~\ref{sec:flywheel} describes the emergence of a community-scale scientific flywheel, including participation as well as evaluation and validation in platform-scale scientific production. Finally, Section~\ref{sec:outlook} offers an outlook on \emph{Science-as-a-Service} and human--AI co-evolution in scientific practice.

%% file: sec-2-bohrium.tex
Section~\ref{sec:intro} argued that the limiting factor for agentic science is no longer model capability alone, but the absence of a scientific stack and shared environments that are explicitly \emph{agent-ready}. We now make this concrete by describing \emph{Bohrium} as infrastructure for scientific production\footnote{\url{https://www.bohrium.com/}}: a platform that turns diverse scientific assets into \emph{governed, callable capabilities} (with traceable execution and explicit constraints) for Reading, Computing, and Experiment, and exposes them to both humans and agents through stable interfaces with execution-level observability.

\subsection{From Bare Scientific Assets to Agent-Ready Capabilities}
\label{subsec:bohrium_bare_to_capabilities}

In many AI4S settings, datasets, computing clusters, laboratory equipment, and scientific software are loosely treated as ``infrastructure''. In practice, they are often \emph{bare assets}: scripts tied to particular clusters, workflows embedded in ad hoc orchestration, data scattered across internal storages, and instruments controlled by bespoke interfaces. These assets may be powerful for human experts, but from an agent's perspective they are opaque, non-uniform, and difficult to call or combine reliably, making them unsuitable for long-horizon agentic workflows.

Bohrium occupies the platform layer between such assets and higher-level orchestrators such as SciMaster. It provides uniform abstraction across cloud and HPC environments, cross-platform scheduling and runtime support for long-running and asynchronous jobs, and a capability registry for tools, models, datasets, and workflows, together with platform-wide logging, monitoring, access control, and accounting. Crucially, Bohrium standardizes previously bespoke assets into \emph{capabilities} with a minimal contract: well-defined inputs/outputs, reproducible execution envelopes, and traceable, auditable runs. As a result, tools, workflows, datasets, and collaboration artifacts can be discovered, scheduled, traced, audited, and cost-accounted for in a consistent way.

This turns Bohrium into a production environment for running scientific workloads across Reading, Computing, and Experiment, while presenting agents with a catalog of callable capabilities with explicit semantics, constraints, costs, and failure modes---a prerequisite for the cycle-running systems discussed in Section~\ref{sec:intro}, where workflows must be executed and compared rather than merely proposed.

\subsection{Reading, Computing, and Experiment as Executable Services}
\label{subsec:bohrium_rce}

\begin{figure}[t]
\centering
\includegraphics[width=\linewidth]{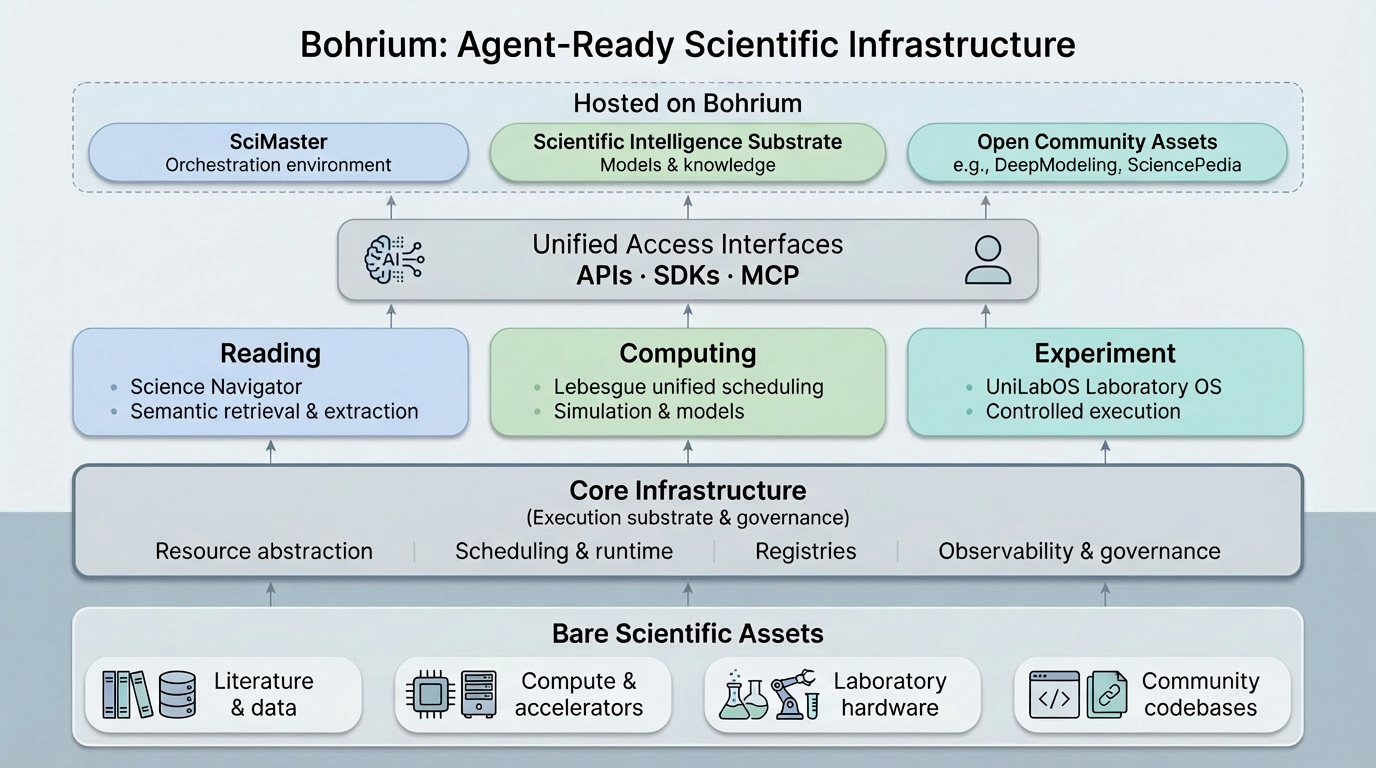}
\caption{\textbf{Bohrium as agent-ready scientific infrastructure.}
Bohrium structures scientific execution around three capability domains---Reading, Computing, and Experiment---and provides a unified execution substrate that makes scientific assets callable, observable, and governable. Through consistent access interfaces for both agents and human users, these capabilities support reliable, tool-augmented scientific workflows spanning evidence access, computation, and experimentation.}
\label{fig:bohrium}
\end{figure}

We next describe three representative subsystems hosted on Bohrium that make the Reading, Computing, and Experiment stages directly accessible through agent-ready capabilities: \emph{Science Navigator} for large-scale literature and patent access, \emph{Lebesgue} for unified computing, and \emph{UniLabOS} for laboratory execution. Figure~\ref{fig:bohrium} overviews how these subsystems expose capabilities through unified interfaces with observable, governed execution.

\textbf{Reading: Science Navigator as an evidence substrate.}
Science Navigator ingests and organizes diverse scientific corpora---research papers, patents, technical reports, and domain-specific repositories---into a unified, AI-ready evidence substrate. Raw documents are normalized, de-duplicated, and enriched with metadata; parsers convert multimodal scientific content into machine-usable structures, including tables and figures, mathematical expressions, chemical formulae and reaction schemes, and domain-specific representations such as molecular structures and spectra when present.

On top of the processed corpus, Science Navigator provides multi-granularity semantic access---from papers and sections down to entities, claims, and workflow fragments. A family of components turns this corpus into a trace-backed evidence substrate: document understanding pipelines extract entities, relations, and experimental or computational workflows; citation and dependency graphs support structured navigation and context expansion; and API layers expose semantic search, multi-hop traversal, and citation-aware exploration. Critically, retrieved passages, extracted facts, and assembled workflow elements are explicitly linked back to their source documents, sections, and citation contexts, enabling provenance-preserving evidence use and replayable inspection and verification. This enables higher-recall, higher-precision scientific search and supports master agents such as \emph{PaSaMaster}, \emph{SurveyMaster}, \emph{PharmMaster}, and \emph{MatMaster} in reading-heavy workflows (Section~\ref{sec:master-agents}). As usage accumulates, access patterns and success/failure signals can refine indexing, ranking, and extraction pipelines, supporting a progressively more reliable evidence substrate over time.

\textbf{Computing: Lebesgue and diverse compute backends.}
On the computing side, Bohrium provides a cross-platform orchestration and execution environment, with the \emph{Lebesgue} scheduling system at its core. Lebesgue abstracts over compute resources---cloud, on-premise HPC, and specialized accelerators---offering a unified interface for job submission, resource allocation, and monitoring. Scientific codes and pipelines are packaged with reproducible environments to run under this scheduler, and their metadata are recorded in registries that capture versions, dependencies, and execution envelopes.

Because Bohrium supports multi-user and multi-agent operation at scale, Lebesgue also provides multi-tenant resource governance mechanisms such as per-job and per-user accounting, quota enforcement, and priority-based scheduling policies. These controls make it possible for many concurrent scientific workflows to share the same compute substrate while maintaining predictable performance and cost boundaries. For agents, Lebesgue turns simulation codes, numerical solvers, and learned surrogates into callable services with unified interfaces and predictable costs. Master agents such as \emph{PDEMaster}, \emph{FlowXMaster}, \emph{PhysMaster}, \emph{OPT-Master}, and \emph{MatMaster} reuse this infrastructure to construct workflows for numerical analysis and design (Section~\ref{sec:master-agents}).

\textbf{Experiment: UniLabOS as an operating system for laboratories.}
For experimentation, Bohrium integrates \emph{UniLabOS}, an operating system for autonomous and semi-autonomous laboratories. UniLabOS connects laboratory hardware (robots, reactors, characterization instruments), online analytics, and scheduling and control logic into a unified execution environment. Experimental procedures---sample preparation, measurement sequences, and instrument operations---are encoded as machine-readable protocols; execution traces, sensor data, and outcomes are logged in a structured form.

By virtualizing instruments and material flows into a transaction-safe substrate, UniLabOS makes experimentation schedulable and auditable through platform interfaces, enabling agents to invoke experiments within predefined envelopes. At the same time, laboratory execution is inherently safety- and reliability-constrained: UniLabOS supports controlled execution policies (e.g., parameter bounds and protocol-level constraints) and human-in-the-loop oversight for higher-risk operations, so that agent-invoked experiments remain auditable and operate within predefined constraints. This allows DMTA-style cycles to be configured, monitored, and analyzed at scale, and enables dry--wet closed loops such as those implemented by \emph{MatMaster} and \emph{AMTechMaster}.

\subsection{Bohrium as an Execution Substrate for Models and Workflows}
\label{subsec:bohrium_models_workflows}

An important design point is that Bohrium is not only a host for ``tools'' in the narrow sense, but an execution substrate on which \emph{models} and \emph{workflows} become first-class, callable, and governable assets. In production scientific systems, models (classical solvers, scientific foundation models, domain surrogates) are invoked repeatedly under varying constraints of cost, latency, reliability, and provenance. Treating them as executable services---rather than as ad hoc libraries bound to a particular runtime---is therefore essential for large-scale agentic science, where orchestrators must route, compare, and reuse alternatives under explicit constraints.

Concretely, Bohrium provides (i) registries that version models and workflows alongside their dependencies, input/output schemas, and execution envelopes; (ii) scheduling and resource governance through Lebesgue so that model and workflow executions are shareable across many concurrent users and agents; and (iii) platform-wide observability so that each invocation can be traced, replayed when needed, audited, and compared across alternatives. This model- and workflow-centric execution layer enables an orchestrator such as SciMaster to route between tools and models, reuse intermediate artifacts, and evaluate outcomes at the workflow level rather than at the single-call level.

This section focuses on the execution and governance properties required for model- and workflow-centric scientific production. The question of how scientific intelligence is organized across a hierarchy of models, and how open communities curate and evolve these assets, is discussed in Section~\ref{sec:intelligence_substrate}.

\subsection{Bohrium as a Living and Evolvable Infrastructure}
\label{subsec:bohrium_living}

The picture that emerges is not a static stack, but a \emph{living infrastructure} that co-evolves with its users, addressing the weak flywheels identified in Section~\ref{subsec:bottlenecks}. Bohrium's logging, tracing, and monitoring facilities record real workloads from both human and agentic users: which capabilities are invoked, how jobs are scheduled, where failures occur, and how resources are consumed. These signals---including failure modes and recovery paths, validation outcomes, and adoption patterns---feed back into scheduling and orchestration policies, onboarding and deprecation of capabilities, and the design of higher-level abstractions exposed to agents.

Equally important, Bohrium is designed to be adaptable as scientific practice evolves: new tools and workflows can be onboarded as standardized capabilities, and external resources can be integrated through the same platform conventions, allowing the substrate to expand and update without fragmenting interfaces. This adaptivity, together with observability and governance, supports long-lived scientific workflows that improve as reusable capabilities accumulate.

At the same time, Bohrium's accounting mechanisms support sustainable operation and long-term maintenance of shared capabilities and laboratory assets. This encourages a shift from one-off research code and isolated experimental setups toward reusable, deployed scientific solutions that remain available to both human scientists and AI agents, and it provides the raw signals on which ecosystem-level incentive and transaction mechanisms can later be explored.

In summary, Bohrium provides the platform layer that turns scientific assets into agent-ready capabilities, exposes Reading, Computing, and Experiment as executable services, and supplies the observability, governance, and evolvability needed for large-scale scientific production. The next section turns to the \emph{scientific intelligence substrate}---how models, knowledge, and communities cohere into a hierarchy of scientific capabilities and how these are reused and evolved at ecosystem scale.

%% file: sec-3-intelligence_substrate.tex
Section~\ref{sec:bohrium} described Bohrium as an execution substrate that makes scientific assets callable, observable, and governable at scale. A natural next question is where \emph{scientific intelligence} resides in such a system. Rather than equating intelligence with a single large model, we argue that in production-scale scientific settings it emerges from a \emph{scientific intelligence substrate}: a combination of (i) a hierarchy of scientific models with complementary roles, (ii) a shared knowledge substrate that makes conceptual and validation structure reusable, and (iii) open communities that curate, maintain, and evaluate capabilities over time.

Throughout this section, we use ``scientific intelligence'' in a system-level sense: not only reasoning and representation, but also the reusable components---models, solvers, workflows, and evaluation artifacts---that make scientific work trace-backed and executable in practice. These components form the interface between agentic orchestration (SciMaster) and governed execution (Bohrium), enabling workflows to be composed, audited, and improved through use within the Bohrium+SciMaster ecosystem.

\subsection{Why and What: A Hierarchy of Scientific Models for Representation, Reasoning, and Execution}
\label{subsec:model_hierarchy}

Scientific tasks differ from generic language tasks in both structure and verification requirements. They routinely involve multimodal scientific artifacts (text, equations, structures, spectra, and logs), require multi-step reasoning with explicit checks, and depend on external computation and experimentation. These properties make it difficult for any single model to provide robust, trace-backed scientific capability across domains and workflows.

Instead, we adopt a hierarchical view in which different classes of models play complementary roles:
(i) \emph{general-purpose scientific foundation models} that focus on task understanding, cross-domain reasoning, and coordination across tools and workflows;
(ii) \emph{domain foundation models} that are grounded in \emph{specific scientific modalities and representations} (e.g., molecular graphs, three-dimensional atomic structures, biological sequences) and encode strong inductive biases through large-scale pretraining (e.g., DPA~\cite{zhang2025dpa3}, Uni-Mol~\cite{zhou2023unimol}, Uni-RNA~\cite{wang2023unirna});
and
(iii) \emph{pipeline- or application-specific models} that are embedded within concrete workflows and optimized for executability, stability, and evaluability under real operating constraints (e.g., multimodal scientific document parsing via Uni-Parser~\cite{fang2024uniparser}, molecular structure--property modeling via Uni-QSAR~\cite{uniqsar2023}, and scientific image representation via Uni-AIMS~\cite{hong2025uniaims}).

This hierarchy reflects both technical and organizational realities. Domain foundation models exploit specialized representations and constraints (e.g., geometric invariances, physical priors, and sequence locality) and are evaluated using modality- and domain-appropriate criteria. General-purpose models emphasize evidence integration, planning, and action routing across heterogeneous capabilities. Pipeline and application models implement workflow-facing transformations and decision logic---such as parsing, alignment, filtering, validation, and adaptor modules---that connect foundation models and tools to repeatable end-to-end pipelines.

The hierarchy becomes operational only when these models are executable and composable on shared infrastructure: they must be invocable through stable interfaces, comparable under controlled execution envelopes, and reusable as workflow building blocks rather than isolated artifacts. In Bohrium+SciMaster, this viewpoint makes model choice and composition an explicit, auditable part of scientific workflows, enabling workflow-level evaluation beyond single-model benchmarks.

\subsection{Innovator: A General-Purpose Scientific Foundation Model}
\label{subsec:innovator}

Within this hierarchy, \emph{Innovator} serves as a general-purpose scientific foundation model that acts as a cognitive hub for scientific tasks. Its role is not to replace domain foundation models or specialized solvers, but to interpret scientific objectives, reason across diverse scientific inputs, and propose executable action plans that can be compiled and carried out through systems such as SciMaster.

\textbf{Scientific perception.}
Innovator is designed to treat scientific inputs as structured, multimodal objects rather than flattened text. It supports the joint interpretation of natural language, mathematical expressions, symbolic representations, molecular and material structures, spectra, figures, and other artifacts commonly encountered in research practice. Many scientific tasks require perception that is informed by domain knowledge---for example, normalizing Markush structures in chemical patents, or jointly reasoning over reaction equations, experimental conditions, tables, and schematic figures in reaction reports. In such settings, perception produces structured representations that can be inspected, compared, and used for downstream scientific actions.

\textbf{Scientific cognition and planning.}
Beyond perception, Innovator emphasizes scientific reasoning, planning, and programming. The central difficulty here is \emph{depth}: many scientific tasks require mechanistic, domain-grounded reasoning and long-horizon problem decomposition akin to expert scientific cognition, with explicit assumptions, alternative hypotheses, and falsifiable checks. Innovator translates high-level scientific questions into structured problem representations, generates hypotheses and solution strategies, and expresses these strategies in executable forms such as code, simulation configurations, or experimental plans. Tasks such as spectral interpretation illustrate iterative cognition: perceptual hypotheses are generated, tested via external tools or databases, revised in light of inconsistencies, and validated against explicit criteria or human judgment. Programming is treated as a primary mechanism for externalizing reasoning and interfacing with computation and experimentation through SciMaster-orchestrated workflows.

\textbf{Tool and capability awareness.}
A defining property of Innovator is its awareness of tools, models, and workflows as first-class entities. The difficulty is \emph{breadth and precision at production scale}: when tens of thousands of tools and models are available, the system must reliably route tasks to appropriate capabilities under explicit constraints, respect interface contracts and operating envelopes (e.g., cost, latency, reliability, and safety), and attach verifiable expectations for downstream execution and validation. In this division of labor, Innovator focuses on scientific reasoning and decision-making, while execution, scheduling, and auditability are handled by SciMaster on top of Bohrium. In this sense, Innovator proposes and updates actions and hypotheses, whereas SciMaster executes and governs them under trace-backed execution.

A stable release of the Innovator model family is under development. In this work, we treat Innovator primarily as a \emph{system role} within the scientific intelligence substrate: its capabilities are realized only when coupled to a governed execution substrate, a standardized tool/capability registry, and trace-backed workflow evaluation, rather than as a claim about a fixed model release, architecture, or benchmark-specific performance.

\subsection{SciencePedia: A Shared Knowledge Substrate for Trace-Backed Reasoning}
\label{subsec:sciencepedia}

Scientific intelligence also depends on structured representations of knowledge that go beyond raw documents and data. While evidence systems such as Science Navigator organize papers, patents, and datasets, \emph{SciencePedia} focuses on scientific \emph{concepts}, assumptions, and reasoning processes, and represents them with explicit provenance and dependency structure.\footnote{See \url{https://sciencepedia.bohrium.com/} for an overview of the SciencePedia project.}

Methodologically, SciencePedia is built by \emph{decompressing} scientific knowledge: instead of storing only compressed conclusions, it synthesizes verifiable \emph{long chains of thought} (LCoT) that make intermediate steps, assumptions, and dependency relations explicit, and then projects this structured reasoning base into an emergent, navigable encyclopedia~\cite{sciencepedia_lcot}. Beyond long-chain reasoning, we are exploring extensions toward trace-backed \emph{long chains of execution}, in which large-scale tool and workflow traces (what was executed, under which constraints, and with what outcomes) are recorded and linked to corresponding concepts and checks, pointing toward a unified \emph{content--tool} knowledge base that supports end-to-end scientific work.

This structured representation transforms static documentation into executable logic. By operationalizing the reasoning components and workflow patterns described above, the knowledge base supports direct inspection and reuse beyond simple text matching. Crucially, this fills a semantic gap in the scientific workflow: it complements evidence retrieval by clarifying the meaning and validity conditions of problem entities, and complements numerical modeling by validating the logical context of what is being simulated.


SciencePedia supports open co-building while remaining auditable: community contributions are organized around explicit provenance and dependency relations, and can be reviewed and iteratively refined using trace-backed validation signals from downstream workflows.

For both humans and AI systems, SciencePedia provides a conceptual substrate that can be queried alongside evidence and computation, enabling workflows that move between literature-grounded evidence, conceptual reasoning, and computational or experimental action. Importantly, SciencePedia is not a single source of truth: it provides structured, provenance-aware representations that facilitate validation and reuse, while leaving empirical checking to the broader evidence, modeling, and experiment layers. In practice, its encyclopedia-style organization across fields and levels encourages cross-domain navigation and comparison, helping surface alternative viewpoints, prerequisite dependencies, and interdisciplinary connections in a more structured and provenance-grounded manner than free-form summaries.

\subsection{DeepModeling: An Open Community Supplying Reusable Scientific Capabilities}
\label{subsec:deepmodeling}

Open communities play a critical role in the scientific intelligence substrate by developing, maintaining, and validating reusable scientific capabilities. \emph{DeepModeling} is an open AI4Science community that develops and curates reusable open-source software, models, workflows, and utilities for scientific computation, with broad coverage across physics, chemistry, and materials research.\footnote{See the DeepModeling community manifesto for its scope and organizing principles: \url{https://github.com/deepmodeling/community/blob/main/MANIFESTO/MANIFESTO.md}.} Its contributions span multiple layers: interoperable engines and solvers, workflow and data utilities for repeatable scientific production, and task-oriented pipelines that package domain methods into reusable building blocks.

At the engine and solver level, representative projects include \emph{DeePMD-kit} for deep potential-based atomic modeling~\cite{WANG2018178}, \emph{ABACUS} for first-principles electronic structure calculations~\cite{abacus}, \emph{DeePTB} for deep learning-based electronic structure modeling~\cite{gu2024deeptb,zhouyinLearning2025}, \emph{dpnegf} for non-equilibrium Green's function (NEGF) quantum transport simulations~\cite{zouDeep2025},  \emph{DeepFlame} for data-driven reactive flow simulation~\cite{mao2023deepflame}, and \emph{jax-fem}, a differentiable finite-element framework built on JAX~\cite{xue2023jax}. Surrounding these engines are workflow and data utilities that support large-scale, repeatable scientific production, such as \emph{DP-GEN}~\cite{zhang2020dpgen}, \emph{dflow}~\cite{liu2024dflow}, \emph{dpdispatcher}~\cite{yuan2025dpdispatcher}, \emph{dftio}~\cite{zhouyinLearning2025} and \emph{dpdata}~\cite{zeng2025dpdata}. At the pipeline and application level, projects such as \emph{APEX}~\cite{li2025apex} and \emph{CrystalFormer}~\cite{taniai2024crystalformer} illustrate how domain methods can be packaged into reusable pipelines and models that can be composed, compared, and iteratively improved over time.

Within the Bohrium+SciMaster setting, these open-source assets can be integrated into governed execution environments and composed into end-to-end workflows alongside other tools, models, and experimental backends. This enables community-developed engines, workflows, and pipelines to be invoked through stable interfaces, scheduled under explicit resource constraints, and assessed via trace-backed workflow outcomes in real scientific workloads.

Beyond individual projects, DeepModeling embodies a long-term approach to building reusable scientific capabilities through open, community-driven development. From its inception, the community has emphasized production-quality software, modular design, and sustained maintenance, reflecting the practical requirements of scientific computation beyond single publications. Rather than converging on a monolithic framework, DeepModeling promotes interoperable components across engines, workflows, and data utilities, allowing contributions at different layers to evolve independently while remaining composable. In this sense, DeepModeling represents an early and sustained experiment in organizing open AI4Science development around long-lived capability building, providing a continuously evolving supply of scientific tools and workflows that can be integrated and exercised within shared execution environments.

In the next section, we describe how SciMaster consumes this scientific intelligence substrate: models provide decision signals, knowledge provides reusable conceptual and validation structure, and open communities supply continuously improving capabilities that can be invoked, composed, and evaluated on shared infrastructure.

%% file: sec-4-scimaster.tex

With Bohrium providing agent-ready capabilities for Reading, Computing, and Experiment, a further question arises: how can these capabilities be organized into coherent, long-horizon scientific workflows carried out by AI co-scientists and AI Scientists? In this section, we present \emph{SciMaster} as an orchestrator and shared environment for scientific AI agents, and describe the federated ecosystem that has formed around it together with ongoing practices for community-in-the-loop evolution.

\subsection{SciMaster as an orchestrator for scientific AI agents}
\label{subsec:scimaster_orchestrator}

SciMaster is an \emph{orchestrator} and shared environment, not a single model\footnote{\url{https://scimaster.bohrium.com}}. It separates reasoning and planning signals (provided by tool-augmented reasoning models and specialized agents) from execution governance on Bohrium, converting scientific problems into structured, tool- and agent-based workflows grounded in Bohrium's execution substrate. For details of the underlying X-Master architecture, its tool-augmented reasoning design, and evaluation results, we refer to Ref.~\cite{chai2025scimaster}.

In production scientific settings, long-horizon workflows are constrained not only by model intelligence, but also by dependency structure, resource boundaries, and verifiability. SciMaster addresses these constraints through system mechanisms that make execution \emph{repeatable under realistic failure modes} and \emph{auditable at workflow granularity}:

\textbf{Task understanding and workflow construction.}
SciMaster maps scientific objectives into executable workflows built from Reading, Computing, and Experiment capabilities. Tool-augmented reasoning models (e.g., X-Master) support task decomposition and planning by interleaving internal reasoning with external tool use through code-based interactions~\cite{chai2025scimaster}. The resulting workflows prioritize executability: dependencies are made explicit (inputs, intermediate artifacts, required capabilities, and invocation contracts), enabling systematic execution rather than ad hoc tool calls. Explicit workflow structure also reduces premature commitment (e.g., treating retrieved evidence as proof, or turning a heuristic plan into a final conclusion) by making intermediate steps inspectable and revisable.

\textbf{Multi-agent coordination and governed scheduling.}
SciMaster coordinates multiple agents with specialized roles (e.g., literature aggregation, numerical modeling, experimental planning) and schedules their actions on Bohrium's capability layer. Coordination is enforced through capability contracts and runtime guards (e.g., schemas, parameter bounds, sandboxed execution, and quota-aware scheduling) rather than assuming perfect agent compliance. This is critical because long-horizon workflows often fail for engineering reasons (resource contention, incompatible interfaces, or merge conflicts across parallel branches) rather than due to a lack of reasoning; SciMaster makes these failure modes observable and controllable.

\textbf{State and memory for long-horizon tasks.}
SciMaster maintains session-level and long-term state for scientific problem entities so that multi-step tasks can unfold over hours or days across many tool calls and agent interactions. This state includes hypotheses, intermediate results, design candidates, and workflow configurations, and is managed as versioned artifacts so that partial results can be reused, compared across branches, and rolled back when needed. This reduces brittleness from implicit state drift, where later steps depend on undocumented intermediate choices that cannot be replayed or debugged.

\textbf{Observability, validation gates, and auditability.}
Tool calls, intermediate artifacts, hypothesis updates, and model invocations are logged with trace metadata to support audit and replay. Workflow-level validation gates determine whether intermediate results and final outputs are accepted, combining interface-level checks (schema and constraint violations) with domain-level checks (scientific consistency, numerical stability, or experimental feasibility). This traceability supports workflow-level evaluation (e.g., cycle time, robustness, and validation outcomes) rather than evaluating nominal autonomy in isolation.

These mechanisms enable characteristic patterns of multi-agent scientific work that are difficult to sustain with tool calls alone. A common workflow skeleton is a staged \emph{Reading--Computing--Experiment} pipeline: literature-grounded hypotheses produce candidate models or parameterizations, simulation-based filtering narrows the search space, and controlled laboratory execution validates or refines candidates. For open-ended problems, SciMaster also supports \emph{branch-and-compare} exploration, where multiple agents propose hypotheses or plans in parallel and results are compared, merged, or discarded based on trace-backed criteria. Human scientists remain in the loop by specifying goals, constraints, and evaluation criteria, while SciMaster provides the runtime that keeps execution governed and replayable.

\begin{figure}[!t]
\centering
\includegraphics[width=\linewidth]{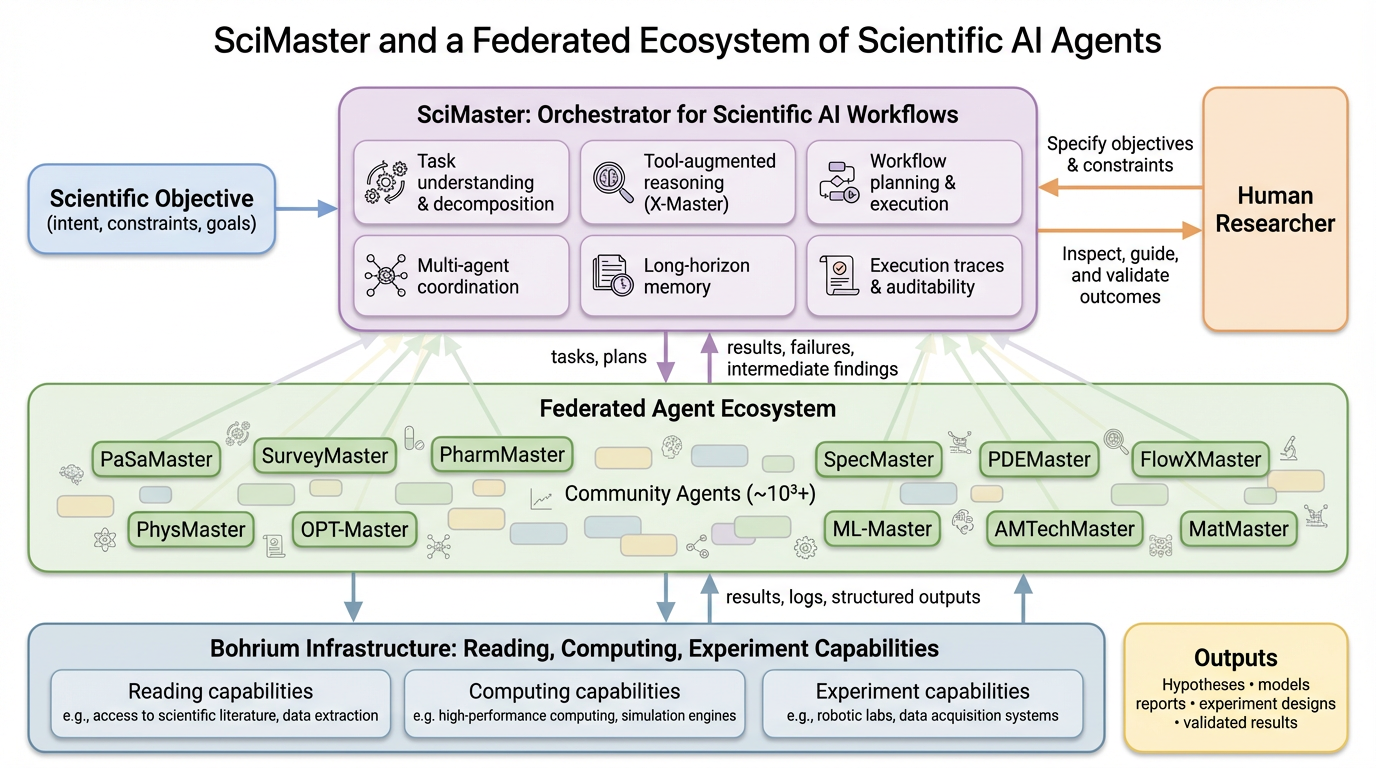}
\caption{
\textbf{Workflow-oriented operation of SciMaster.}
SciMaster acts as a shared orchestration runtime that translates scientific objectives into executable, long-horizon workflows by coordinating agents and invoking Reading, Computing, and Experiment capabilities on Bohrium. Through explicit state management and execution traces, it supports inspectable, auditable, and reusable scientific work across a federated agent ecosystem.
}
\label{fig:scimaster}
\end{figure}

Figure~\ref{fig:scimaster} illustrates this workflow-oriented operation of SciMaster over Bohrium's capabilities.

\subsection{A federated ecosystem of scientific AI agents}
\label{subsec:federated_ecosystem}

SciMaster operates within---and helps shape---a federated ecosystem of scientific AI agents, tools, and contributors. The ecosystem is diverse in both domain coverage and task specialization. In current deployments and community channels, scientific agents and applications are developed at the order of $10^3$ scale across life sciences, chemistry and materials, physics, computational science, mathematics, and cross-domain settings. Their tasks span literature and data analysis, hypothesis generation, simulation and optimization, data analysis and visualization, and experimental planning, forming a practical division of labor rather than a rigid taxonomy.

Federation is enabled not by shared ownership but by shared contracts. Capabilities are packaged with stable schemas, execution envelopes, and trace provenance (and, where applicable, validation checks), so independently developed tools can interoperate within shared workflows. Adoption of Bohrium's packaging conventions reduces friction in onboarding new capabilities and helps keep interfaces consistent as the ecosystem grows. In this setting, independently developed agents can route to shared capabilities, share intermediate artifacts, and participate in comparable workflow evaluation because execution and traces are grounded on the same substrate.

Equally important is the growing developer community around SciMaster, with contributions at the order of $10^3$ scale spanning agents, tools, workflows, and examples through developer tooling and feedback channels. Beyond enabling agents to run on a shared substrate, these channels provide practical pathways for capabilities and workflows to be built, standardized, and onboarded into the ecosystem. Community events facilitate cross-domain exchange and rapid iteration, helping establish SciMaster as a shared environment rather than a closed system.

Within this environment, ``master agents'' are more mature and widely reused workflows, not a privileged class of agent; they coexist with many other community agents on the same substrate and are distinguished mainly by their reliability and impact on real workflows. In Section~\ref{sec:master-agents}, we examine representative master agents in detail.

As Bohrium's accounting and usage signals accumulate, the ecosystem can also support transaction loops grounded in observed usage and outcomes. In the current deployment, usage-based charging and subscription plans provide operational support for shared services and capabilities, while additional mechanisms for linking value creation with rewards for community-developed tools, models, and laboratory capabilities remain an active area of exploration (Section~\ref{sec:flywheel}).

\subsection{Ongoing practices and community-driven evolution}
\label{subsec:community_evolution}

Beyond its current orchestration functions, SciMaster has become a site of active experimentation. We highlight several ongoing practices that illustrate how the ecosystem may evolve as more tools, agents, and workflows accumulate.

\textbf{Agent-assisted capability onboarding.}
Much scientific software lacks interfaces that are stable enough for reliable tool invocation. To reduce the gap between a working research script and an agent-callable capability, we are developing agent-assisted workflows that analyze existing code, propose input--output schemas and invocation contracts, construct reproducible execution environments, and generate invocation examples together with basic validation checks. The goal is not full automation, but lowering friction while preserving correctness and reproducibility; in practice, semantic I/O specification and deterministic execution environments remain key bottlenecks.

\textbf{Trajectory-guided improvement.}
SciMaster's daily operation generates execution trajectories, including tool invocation sequences, workflow structures, failure modes, and correction paths. We are exploring how such trajectories can inform routing decisions, suggest reusable workflow templates, and improve robustness in long-horizon scientific tasks. These efforts remain exploratory: signals are noisy, credit assignment across tools is non-trivial, and aggregation must account for domain shift and differences in user intent.

\textbf{Scientific foundation models and adaptor models.}
Together with the community, we are exploring scientific foundation models that integrate reading, reasoning, and tool use across domains, complementing the X-Master architecture described in Ref.~\cite{chai2025scimaster}. In parallel, we are investigating lightweight adaptor models that capture per-agent preferences and domain specialization to improve stability and domain-appropriateness of tool use, while remaining compatible with SciMaster's orchestration and governance mechanisms. These are treated as optional accelerators rather than requirements, balancing specialization with policy stability across evolving toolchains.

Taken together, these practices suggest that SciMaster is not a fixed system, but an open environment in which infrastructure, algorithms, community contributions, and real scientific problems co-evolve. Rather than converging toward a single AI Scientist, the ecosystem is developing into a setting in which scientific intelligence emerges from coordinated, tool-grounded, multi-agent activity on top of a shared scientific substrate.

A compact one-page card for each master agent is provided in Appendix~\ref{app:agent-cards}.

%% file: sec-5-casestudy.tex
\newcolumntype{L}[1]{>{\raggedright\arraybackslash}p{#1}}

Sections~\ref{sec:intro}--\ref{sec:scimaster} introduced Bohrium as agent-ready infrastructure and SciMaster as an orchestration environment for tool-augmented Reading, Computing, and Experiment. We now examine eleven representative \emph{master agents} built on top of this stack. Here, ``master agents'' do not denote a privileged class or a fixed taxonomy; rather, they refer to agentic workflows that have become relatively mature, reliable, and widely reused within particular domains or application scenarios. Together, they illustrate how diverse scientific tasks---from literature analysis to numerical simulation, optimization, materials design, and experiment-facing execution---can be expressed as coordinated, multi-step workflows once the underlying substrate provides governed access to corpora, models, solvers, and laboratory resources (see Appendix~\ref{app:agent-cards} for one-page cards).

The goal of this section is not to catalog behaviors exhaustively, but to highlight (i) the diversity of scientific problems addressed, (ii) the structural regularities that recur across domains, and (iii) representative reductions in end-to-end scientific cycle time enabled by workflow-level execution and evaluation.

\subsection{Overview across tasks, domains, and maturities}
\label{subsec:master_overview}

Table~\ref{tab:master-overview} summarizes the eleven master agents currently deployed or prototyped within the Bohrium+SciMaster ecosystem. The agents differ in domain logic, maturity, and user-facing interfaces, yet share a common reliance on Bohrium's foundational services (Science Navigator--scale corpora, spectral and structural databases, numerical backends, design-of-experiments modules, and laboratory operating systems) and on SciMaster for orchestrating long-horizon workflows. This shared substrate makes workflow outcomes comparable through trace-backed execution, even when the underlying tools and domain procedures vary substantially.

\begin{table*}[t]
\centering
\scriptsize
\setlength{\tabcolsep}{3pt}
\caption{Overview of eleven master agents and representative reductions in scientific cycle time.}
\label{tab:master-overview}
\begin{tabular}{L{2.2cm}L{2.2cm}L{1.3cm}L{3.2cm}L{4.8cm}}
\hline
Agent & Domain & Maturity & Typical scenario & Efficiency gain (baseline $\rightarrow$ agent) \\
\hline
AMTechMaster &
Additive manufacturing &
Proto. &
Simulation-driven additive manufacturing process parameter design. &
Manual parameter search with sparse simulation feedback $\rightarrow$ automated multi-iteration parameter search guided by simulation. \\
\hline
FlowXMaster &
CFD \& engineering simulation &
Proto. &
From geometry or sketches to meshing, solver configuration, and flow simulation. &
Manual CAD/mesh/solver setup across tools $\rightarrow$ consolidated agent pipeline with substantially reduced configuration overhead. \\
\hline
MatMaster &
Electrolyte materials design &
Prod. &
Closed-loop formulation optimization with computation and experiment. &
Faster screening and fewer invalid experiments in closed-loop runs; formulation cycle shortened substantially (Table values are representative). \\
\hline
ML-Master &
ML experiment automation &
Proto. &
End-to-end machine learning experimentation under fixed time budgets. &
Multi-day expert experiment design/execution $\rightarrow$ competitive performance within a single 12~h agent loop. \\
\hline
OPT-Master &
OR \& optimization &
Proto. &
Natural-language optimization tasks and NP-hard benchmark problems. &
Baseline LLM + heuristics $\rightarrow$ $\sim40\%$ absolute accuracy gain and state-of-the-art circle-packing scores. \\
\hline
PaSaMaster &
Lit.\ search (cross-domain) &
Prod. &
Open-ended, natural-language literature search and project scoping across multiple scientific disciplines. &
Manual multi-engine search ($\sim$1 PhD, 2--3 h) $\rightarrow$ automated high-recall retrieval in $\sim$3 min. \\
\hline
PDEMaster &
Text-to-PDE simulation &
Proto. &
From natural-language partial differential equation (PDE) descriptions to valid numerical simulations. &
Manual derivation + coding (multi-day) $\rightarrow$ automated weak-form derivation and finite-element simulation in $<1$ hour. \\
\hline
PharmMaster &
Med.\ chem.\ \& patents &
Prod. &
Target landscaping and freedom-to-operate (FTO) analysis for candidate molecules. &
Patent landscape study $\sim10$ days $\rightarrow$ $<1$ day; per-molecule FTO assessment $\sim2$ days $\rightarrow$ $\sim10$ min (low false-positive rate in internal evaluations); Markush recognition $\approx90\%$. \\
\hline
PhysMaster &
Theor.\ \& comp.\ physics &
Near-prod. &
Analytical and numerical analysis in high-energy physics, condensed matter, and astrophysics. &
Month-scale lattice QCD data-analysis workflow $\rightarrow$ $\sim$1 day; large fractions of fitting and parameter scans automated. \\
\hline
SpecMaster &
Spectral struct.\ elucidation &
Proto. &
Multi-spectral (NMR, MS, IR) structure elucidation and candidate ranking. &
Day-scale expert interpretation $\rightarrow$ minutes-level spectrum--structure loop; $>50\%$ top-1 accuracy on NMRexp. \\
\hline
SurveyMaster &
Survey writing (cross-domain) &
Prod. &
Preparation of method- or field-level scientific surveys grounded in large corpora. &
Human effort: $1$--$2$ PhDs, $\geq 1$ month $\rightarrow$ structured 40--80 page draft in $\sim4$ h (from $\sim10^3$ papers). \\
\hline
\end{tabular}
\end{table*}

\subsection{Recurring workflow structure across domains}
\label{subsec:master_regularities}

Although these agents span different scientific domains, common workflow patterns emerge once tasks are expressed through SciMaster's orchestration and Bohrium's reusable capability blocks. Table~\ref{tab:master-rce} summarizes typical capabilities and tool types.

\textbf{(1) Multi-stage workflows with explicit artifacts and checks.}
Each master agent decomposes a task into a sequence of stages---interpretation and grounding, tool invocation, verification, and aggregation---with explicit intermediate artifacts and trace-backed validation gates. For example, \emph{PharmMaster} parses patents, reconstructs Markush scopes, builds scaffold networks, and applies similarity search and SAR synthesis before producing an evidence-grounded FTO assessment. \emph{FlowXMaster} interprets geometry or sketches, constructs meshes, configures CFD solvers, executes simulations, and synthesizes outputs. Across domains, these workflows follow a recurring ``interpret--invoke--verify'' pattern: \emph{PDEMaster} checks boundary-condition consistency before launching FEM jobs, while \emph{PharmMaster} filters patent-extracted structures through chemistry validity checks before downstream matching.

\textbf{(2) Reuse of capability blocks across agents.}
The master agents draw from a shared pool of capability blocks exposed on Bohrium. \emph{PDEMaster} and \emph{FlowXMaster}, for instance, both rely on chains of geometry preprocessing, meshing, and solver orchestration, differing mainly in problem specification and downstream analyses. On the molecular side, \emph{PharmMaster} and \emph{MatMaster} reuse common primitives for molecular modeling and property evaluation while targeting different decision problems. Literature-centered agents such as \emph{PaSaMaster}, \emph{SurveyMaster}, and \emph{PharmMaster} all depend on Science Navigator--scale corpora, and their structured outputs can seed modeling, optimization, and experiment-facing workflows.

\textbf{(3) Orchestration as the unifying mechanism.}
SciMaster coordinates capability blocks into coherent workflows through task decomposition, routing, progress monitoring, and failure recovery under explicit contracts and validation gates. In \emph{ML-Master}~\cite{liu2025ml}, SciMaster organizes planning, coding, and execution agents under fixed time budgets. \emph{OPT-Master} leverages SciMaster to manage large-scale search across solvers and heuristic engines. \emph{MatMaster} couples literature extraction, high-throughput computation, design-of-experiments routines, and UniLabOS execution into full ``read--compute--experiment'' cycles. Across domains, orchestration supports robust execution while preserving domain-specific logic through reusable workflow skeletons and guarded tool use.

\textbf{Mini-case: FlowXMaster---end-to-end CFD workflows.}
Traditional CFD campaigns require multiple specialists and tools to move from CAD or sketches through geometry cleaning, meshing, solver configuration, simulation, and post-processing \cite{fan2025chatcfd}. FlowXMaster compresses this process into an agent-driven pipeline: starting from natural-language descriptions or sketches, it reconstructs or refines geometry, generates meshes, configures appropriate solvers, runs simulations via Bohrium's compute layer, and aggregates key indicators into reports consumable by engineers. The hardest steps are typically meshing and stable solver configuration; FlowXMaster mitigates these through interpret--invoke--verify loops with guardrails such as mesh validity checks and template-backed fallback configurations when solver runs fail.

\textbf{Mini-case: ML-Master---time-budgeted machine learning experimentation.}
ML-Master highlights orchestration challenges in computational workflows where search, coding, and evaluation must be balanced under fixed time budgets. Given a task specification and a wall-clock constraint, ML-Master decomposes the problem into planning, code generation, training, evaluation, and refinement phases, and allocates resources dynamically across these phases. SciMaster schedules parallel runs, monitors intermediate performance signals, and triggers fallback strategies when runs fail or underperform. Execution traces---including failed experiments, partial results, and resource usage---are retained and reused to guide subsequent decisions within the same workflow. Rather than optimizing a single model invocation, ML-Master optimizes the \emph{workflow outcome} under time and resource constraints.

\begin{table*}[t]
\centering
\scriptsize
\setlength{\tabcolsep}{3pt}
\caption{Roles of eleven master agents in the Bohrium+SciMaster ecosystem.}
\label{tab:master-rce}
\begin{tabular}{L{2.2cm} L{6.2cm} L{4.8cm}}
\hline
Agent & Typical capabilities & Typical tool types (via Bohrium+SciMaster) \\
\hline
AMTechMaster &
Geometry-to-mesh; thermo-mechanical AM simulation; parameter sweeps and ranking; toward closed-loop AM control. &
Meshing; FEM/JAX-FEM; result analysis; optimization routines; printer and inspection interfaces. \\
\hline
FlowXMaster &
Geometry ingestion and cleanup; meshing; CFD solving; post-processing and comparison; toward dry--wet CFD loops. &
CAD/mesh tools; CFD solvers; visualization and post-processing; workflow engines; sensor interfaces. \\
\hline
MatMaster &
Literature mining; high-throughput computation; design-of-experiments; lab execution; data re-ingestion for closed-loop optimization. &
Scientific databases; DFT/MD engines; ML property predictors; DoE tools; lab operating system; ELN / Bohrium DB. \\
\hline
ML-Master &
Task understanding; pipeline design; performance-oriented code generation and iterative optimization; experiment scheduling, tracking, and failure recovery. &
Code execution environment; training scheduler; experiment trackers, dashboards, and logging tools. \\
\hline
OPT-Master &
Natural-language-to-model translation; solver orchestration; large-scale heuristic and tree search on optimization benchmarks and real tasks. &
MIP/LP solvers; heuristic and RL libraries; parallel search runtime on clustered compute. \\
\hline
PaSaMaster &
Multi-disciplinary natural-language retrieval; multi-stage filtering and ranking; citation expansion and multi-hop exploration; structured evidence collection. &
Open-world literature search; domain indices and databases; ranking and clustering services; citation graph traversal. \\
\hline
PDEMaster &
Parsing PDEs and boundary conditions from text; weak-form derivation; finite-element simulation with self-checking and diagnostics. &
Symbolic tools; meshing; FEM solver stack; job orchestration and logging on Bohrium. \\
\hline
PharmMaster &
Patent parsing; Markush / R-group extraction; scaffold-network construction; SAR synthesis; FTO and risk analysis for candidate molecules. &
Patent/PDF parsers; chemical structure databases; similarity search; SAR analytics; reporting and citation grounding. \\
\hline
PhysMaster &
Task clarification and constraint extraction; precise literature retrieval; long-chain theoretical reasoning; numerical computation and parameter scans. &
Advanced mathematical tools; numerical methods (e.g., QMC/DMRG); optimization libraries; literature search; physics knowledge base. \\
\hline
SpecMaster &
Spectral and text ingestion; candidate generation; multi-spectra consistency checks; integration with characterization workflows. &
Spectrum parsers; spectral/structural databases; forward spectral simulators; rule-based validators; instrument interfaces. \\
\hline
SurveyMaster &
Topic decomposition; corpus construction; outline generation; section-level drafting with citation grounding and consistency checks. &
Literature and citation databases; information extraction and clustering; long-form generation with provenance tracking. \\
\hline
\end{tabular}
\end{table*}

\subsection{Implications for a scientific-agent ecosystem}
\label{subsec:master_implications}

These master agents illustrate how a scientific-agent ecosystem can emerge once foundational capabilities are exposed as reusable and composable building blocks.

First, agents benefit from cross-workflow reuse at two levels. At the artifact level, structured literature sets, extracted tables, candidate lists, meshes, simulation configurations, and experimental protocols can be passed between workflows. At the capability level, solver wrappers, validators, scheduling templates, and evaluation routines can be reused across domains once exposed on a shared substrate. This allows literature-focused agents (\emph{PaSaMaster}, \emph{SurveyMaster}, \emph{PharmMaster}) to seed modeling and analysis agents such as \emph{PDEMaster} and \emph{PhysMaster}, and enables optimization-heavy pipelines (\emph{OPT-Master}, \emph{MatMaster}) to reuse numerical and molecular modules developed elsewhere.

Second, the examples demonstrate extensibility. Agents such as \emph{FlowXMaster} and \emph{AMTechMaster} interface with measurement systems and move toward dry--wet loops, while abstractions developed for optimization and control can be repurposed across domains without reimplementing core execution logic.

Finally, the reported efficiency gains indicate that agent-ready scientific infrastructure can deliver substantial practical benefits in representative settings. Across the eleven agents we observe reductions from months to days in numerical physics and from weeks to hours in literature-driven workflows, alongside substantial shortening of closed-loop materials optimization cycles (Table~\ref{tab:master-overview}). Together, these case studies suggest that Bohrium+SciMaster supports not only individual agents, but also a growing, interconnected ecosystem whose effectiveness is judged by sustained impact on real workflows rather than nominal autonomy in isolation.

%% file: sec-6-flywheel.tex
Taken together, the agent-ready execution substrate described in Section~\ref{sec:bohrium}, the scientific intelligence substrate introduced in Section~\ref{sec:intelligence_substrate}, and the empirical evidence from the master-agent case studies in Section~\ref{sec:master-agents} point to a broader structural pattern that is beginning to take shape. Crucially, this pattern is not centered on any single ``AI Scientist'' model, but on a shared, agent-ready platform on which tools, models, knowledge artifacts, agents, and human researchers coexist and interact under trace-backed execution. Rather than functioning as isolated components, these elements increasingly form a coupled process that organizes \emph{scientific production} through repeated cycles of execution, evaluation, and improvement. We refer to this coupled pattern as a \emph{community-scale scientific flywheel}.

\begin{figure}[t]
  \centering
  \includegraphics[width=\textwidth]{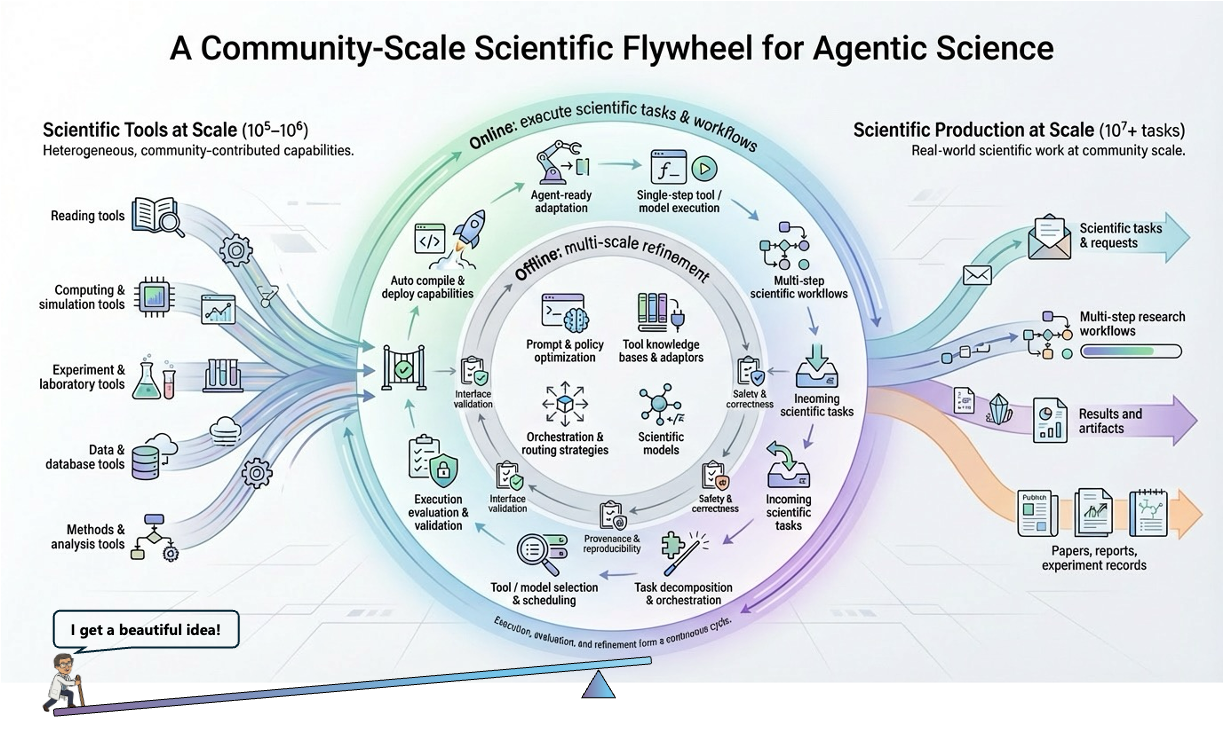}
  \caption{
  \textbf{The emergence of a community-scale scientific flywheel.}
  At platform scale (a multi-million scientific user base and large collections of tools, workflows, and models), scientific production can be organized by a coupled online--offline loop. The outer loop captures online execution on real tasks under realistic constraints and the resulting trace/validation signals, while the inner loop captures offline refinement (e.g., packaging, validation gates, routing policies, and model/knowledge updates) that feeds improvements back into subsequent execution.
  }
  \label{fig:flywheel}
  \vspace{-1.0em}
\end{figure}

\subsection{Forward and Backward Propagation in Scientific Production}
\label{subsec:flywheel_forward_backward}

At the core of the flywheel is a coupling between online execution and offline refinement (Figure~\ref{fig:flywheel}). Online, capabilities are invoked at scale to solve real tasks under realistic resource and safety constraints. Offline, the resulting execution traces and validation outcomes are consolidated into updates that improve subsequent execution. This coupling makes end-to-end scientific production more \emph{observable, comparable, and continuously improvable} across diverse tools and backends, rather than treating each tool, model, or pipeline as a standalone artifact.

In the \emph{forward} direction, reusable capabilities---tools, workflows, models, and knowledge components---are integrated into the shared substrate and exposed for governed execution, so that agents and human researchers can invoke them reliably within long-horizon workflows. Capabilities are first exercised through atomic invocations that establish basic interface checks and trace patterns. They are then composed into multi-step workflows that reflect real research practice. Executed by both human researchers and AI agents, these workflows produce not only scientific outcomes, but also structured traces: task decompositions, tool and model invocations, intermediate artifacts, validation checkpoints, and failure modes.

In the \emph{backward} direction, traces and validation outcomes become first-class signals for system-level refinement. Workflow performance can be assessed along multiple dimensions, including correctness and constraint satisfaction, replayability and provenance, success and failure patterns, cost and latency, and robustness under realistic constraints. When available, feedback can propagate backward across the stack: updating capability packaging and interface contracts, strengthening validation gates and safety envelopes, improving orchestration templates and routing strategies, and refining curated knowledge artifacts. When appropriate, such signals may also support offline improvement of general-purpose and domain models by grounding learning and evaluation in traces of real scientific problem solving rather than benchmark-only performance.

A defining property of this flywheel is the separation---and coupling---between online execution and offline refinement. Online execution must handle scale and diversity of tasks. Offline processes aggregate signals across time, domains, and workflows. Some updates are fast-moving (e.g., orchestration templates, routing policies, interface constraints, and validation gates). Others are slower-moving (e.g., curated knowledge bases, maturation of domain capabilities, and model refinement). Together, these processes enable cumulative improvement that is difficult to achieve through manual engineering or isolated demonstrations.

\subsection{How to Participate: Co-building Infrastructure and Ecosystems}
\label{subsec:flywheel_participation}

A community-scale flywheel cannot be built by any single team. Modern scientific production is too diverse, and the long tail of tools, data, and experimental settings is too broad. The practical requirement is therefore co-building: shared infrastructure that lowers the cost of making capabilities agent-ready, and shared ecosystem processes that make contributions reusable, traceable, and evaluable.

Participation can take different forms, with different degrees of openness: from fully open-source releases, to shareable benchmarks and workflow templates, to private executions that expose only minimal, aggregated signals. The common requirement is that contributions enter the flywheel through shared contracts and traces, so they can be deployed, assessed, reused, and iterated on a common substrate.

\textbf{Infrastructure and capability contributors.}
For teams that maintain scientific tools, datasets, or workflows, Bohrium offers a path from ``working code'' to governed, reusable capabilities: clarified inputs/outputs, reproducible execution envelopes, and basic validation checks. In return, onboarded capabilities can be invoked across many tasks and users, reducing maintenance burden through standardized execution, improving reliability through trace-backed debugging, and enabling comparisons against alternative implementations under common runtime and cost constraints.

\textbf{Open AI4S communities and maintainers.}
Open communities provide durable capability building blocks spanning engines, solvers, workflows, models, benchmarks, and best practices (e.g., DeepModeling). When integrated into shared execution and traceability infrastructure, these assets become easier to invoke, compose, and maintain at scale: packaged execution reduces ``it works on my cluster'' friction, platform-scale use produces concrete bug reports and performance envelopes, and trace-backed assessment improves discovery, reuse, and iterative improvement.

\textbf{Knowledge contributors and educators.}
Scientific platforms also benefit from contributors who structure and explain knowledge. \emph{SciencePedia} supports collaborative construction of conceptual structures, assumptions, and derivations with explicit provenance. SciencePedia supports open co-building while remaining auditable: contributions are organized around explicit provenance and dependency relations, and can be reviewed and iteratively refined using trace-backed signals from downstream workflows. Participation includes authoring new entries, improving clarity, adding references, curating reading paths, and contributing worked examples that connect concepts to evidence queries and reusable workflow patterns. Provenance and dependency structure make it easier to audit which assumptions a workflow relies on when results are later validated or challenged.

\textbf{Agent and workflow builders.}
For developers who build agents on top of SciMaster, participation turns tacit know-how into explicit, replayable workflow structures. When agents are executed on real tasks, their trace-backed runs make interface gaps, failure modes, and recurrent workflow structures explicit. Contributors benefit because these traces make debugging and regression testing practical, shared capability catalogs reduce integration cost, and reusable templates shorten the time from a new task to a working, robust pipeline.

\textbf{Scientific practitioners and real-task users.}
Finally, scientific practitioners running real tasks provide the grounding signals that keep the flywheel tied to practice. Participation does not require publishing proprietary data or unreleased results: in many cases the most useful signals are operational and minimal (which workflow succeeded or failed, which constraints were binding, which steps required human intervention, and how cost and latency traded off against quality). In return, practitioners benefit from improved reliability of commonly used workflows, faster iteration through reusable capability blocks, and clearer performance envelopes that reduce the risk of silent failures in long-horizon agentic work.

\subsection{Rethinking Evaluation and Validation for Platform-Scale Scientific Production}
\label{subsec:flywheel_eval_validation}

As scientific production becomes platform-scale---that is, executable and trace-backed, and continuously refined through use---evaluation and validation shift from peripheral concerns to central organizing mechanisms. In such systems, evaluation and validation are not merely post hoc judgments, but processes that govern reusable assets, coordinate contributors, and shape how scientific value is created and recognized.

A first shift concerns \emph{what} is evaluated and validated. In traditional scientific practice, the paper is a primary unit of value assessed through discrete peer review. In platform-scale scientific production, the units of value expand to include executable workflows, validated tools and models, curated knowledge artifacts, and their observable impact on real tasks under realistic constraints. Validation emphasizes correctness, consistency with physical or experimental constraints, and replayability where applicable. Evaluation emphasizes efficiency, robustness, reuse, and long-term contribution to the ecosystem.

A second shift concerns \emph{how} evaluation and validation are implemented. Rather than relying on a single global metric, signals are embedded throughout the stack. Capabilities are validated at interfaces through schemas, constraints, and safety checks. Models are validated through domain-specific tests and consistency criteria. Workflows and agents are evaluated end-to-end based on success rates, cost, latency, and robustness. Knowledge artifacts are evaluated through provenance tracking and explicit dependency structure. This layered approach helps validation safeguard scientific correctness while evaluation guides system-level optimization and evolution.

Finally, these changes interact with the publication and peer-review pipeline. Peer review remains essential for interpretation, synthesis, and claims of novelty, but it is no longer the sole mechanism for establishing trust at scale. Continuous, usage-grounded signals---such as adoption, replayability, validated impact, and trace-backed failure analysis---provide complementary evidence about reliability and contribution. Over time, this combination supports a more pluralistic evaluation regime, in which papers, executable artifacts, and community-curated capabilities coexist as recognized outputs of scientific activity.

Together, these shifts indicate that the scientific flywheel described above is not only a technical feedback loop, but also a re-organization of scientific production itself. With execution, participation, evaluation, and validation tightly coupled on a shared substrate, the pace of iteration can increase while remaining grounded in checked practice. In the next section, we turn to broader implications for \emph{Science-as-a-Service} and the long-term co-evolution of humans and AI systems in scientific work.

%% file: sec-7-conclusion.tex
This work has presented Bohrium+SciMaster not merely as individual systems, but as a coupled \emph{infrastructure-and-ecosystem} for \emph{agentic science at scale}. Bohrium provides an execution substrate that turns diverse scientific assets into agent-ready capabilities for Reading, Computing, and Experiment, while SciMaster organizes these capabilities into long-horizon, tool-augmented scientific workflows. Together with a scientific intelligence substrate---model hierarchies, shared knowledge substrates, and open communities that curate reusable capabilities---this stack supports scientific work as an executable and observable process whose improvement is grounded in trace-backed execution and governed interfaces.

The eleven master agents examined in Section~\ref{sec:master-agents} illustrate how this approach changes the unit of scientific automation. Rather than treating tools or models as isolated components, representative tasks are expressed as coherent, multi-step workflows that integrate evidence use, computation, optimization, and experiment-facing validation. Across literature-driven analysis, numerical simulation, and dry--wet closed-loop optimization, the system yields measurable reductions in end-to-end cycle time (Table~\ref{tab:master-overview}). Importantly, these gains arise not from any single model or algorithm, but from infrastructure-level design choices that make execution callable, composable, and evaluable at workflow granularity.

A central implication is that agentic science enters a different operating regime once it runs on shared, production-grade environments. At multi-million scale, execution traces, validation outcomes, and reuse signals accumulate across tasks and domains, making it possible to iteratively refine capability packaging, orchestration templates, and model/workflow choices based on observed performance, while maintaining auditability and human oversight. As more users move from bespoke scripts and fragmented pipelines to agent-ready services, the system increasingly functions as a scientific platform in which content (papers, data, protocols), capabilities (models, solvers, laboratory modules), and actors (human researchers and AI agents) are co-located on a shared substrate and connected through trace-backed execution.

This infrastructure- and ecosystem-centric perspective contrasts with approaches that center scientific automation on a single end-to-end agent. Instead, Bohrium+SciMaster supports a spectrum of \emph{scientific agents}---from single-capability tools and workflow-specific agents, to collaborative ``co-scientist'' agents that assist human researchers, and potentially to more autonomous \emph{AI Scientist}-style agents. By grounding these agents in a shared substrate with governed interfaces and trace-backed execution, the platform enables coexistence, reuse, and sustained improvement across levels of autonomy.

Looking forward, this framing suggests a natural path of evolution. As more scientific capabilities become standardized and agent-ready, the cost of building domain agents and workflows can drop substantially. At ecosystem scale, accumulated traces, validation outcomes, and human feedback can, when appropriate, drive offline refinement of orchestration policies, workflow templates, interface constraints, and scientific models, grounding system evolution in real workloads rather than static benchmarks.

More broadly, these developments point toward \emph{Science-as-a-Service} as an organizational paradigm. In this view, humans and AI agents collaborate on a common substrate: humans provide intent, interpretation, and accountability, while agents execute and iterate under governed interfaces. Shorter iteration cycles alone do not guarantee better science; rather, the value of such systems lies in enabling progress that remains auditable, responsibly governed, and centered on scientific integrity and human judgment.

%% file: sec-append.tex
In this appendix, we provide one-page cards for the eleven master agents introduced in Section~\ref{sec:master-agents}, summarizing their identity, workflows, underlying tool stacks, efficiency gains, and representative case studies. Currently, MatMaster and SpecMaster are available for online evaluation via the SciMaster homepage; the remaining master agents are scheduled for sequential deployment beginning in early 2026.

\clearpage
\begin{agentcard}[AMTechMaster]

\textbf{Identity}

\agentfield{Name}{AMTechMaster}
\agentfield{Domain}{Additive manufacturing}
\agentfield{Maturity}{Prototype}
\agentfield{Tagline}{Simulation-driven process optimization for metal additive manufacturing.}

\vspace{0.35em}
\textbf{Problem and workflow}

\textit{Traditional bottleneck.}
Metal additive manufacturing process design involves complex, tightly coupled
steps spanning geometry definition, meshing, multi-physics simulation, and
parameter tuning. These workflows require substantial expert intervention and
repeated trial-and-error, resulting in long iteration cycles and limited
exploration of feasible process windows.

\textit{Agentic workflow.}
AMTechMaster executes an end-to-end simulation-driven optimization loop. Given
natural-language descriptions of material, geometry, and process constraints, it
automatically constructs geometries and meshes, configures multi-physics
simulations, analyzes defect-related signals such as residual stress, and
iteratively refines process parameters to converge on optimal printing
conditions.

\vspace{0.35em}
\textbf{Underlying stack and execution model}

AMTechMaster operates as a simulation-centric manufacturing agent on Bohrium:
\begin{itemize}[leftmargin=*]
  \item \emph{Automated pre-processing:} geometry construction and meshing from
        structured or natural-language inputs.
  \item \emph{High-fidelity simulation:} thermo-mechanical finite-element
        simulation of metal printing processes.
  \item \emph{Iterative optimization:} result-driven parameter updates to guide
        convergence toward low-defect process windows.
\end{itemize}

\vspace{0.35em}
\textbf{Efficiency and impact}

AMTechMaster significantly reduces manual setup and expert intervention in
additive manufacturing simulations, enabling faster convergence toward optimal
process parameters and lowering the barrier to simulation-informed process
design.

\vspace{0.35em}
\textbf{Representative case}

\textit{Residual-stress minimization in metal printing.}
For an IN718 single-track printing task under realistic power constraints,
AMTechMaster iteratively explored the process window via coupled
thermo-mechanical simulation and identified printing parameters that minimize
predicted residual stress while remaining compatible with practical
manufacturing conditions.

\end{agentcard}

\clearpage
\begin{agentcard}[FlowXMaster]

\textbf{Identity}

\agentfield{Name}{FlowXMaster}
\agentfield{Domain}{CFD and engineering simulation}
\agentfield{Maturity}{Prototype}
\agentfield{Tagline}{Language or images $\rightarrow$ executable CFD workflows.}

\vspace{0.4em}
\textbf{Problem and workflow}

\textit{Traditional bottleneck.}
High-fidelity CFD spans geometry, meshing, solver setup, simulation, and post-processing across fragmented tools. The workflow is expert-intensive and manual, making CFD slow to use for rapid design iteration.

\textit{Agentic workflow.}
From a short description or an input image, FlowXMaster reconstructs or generates 3D geometry, builds meshes, configures solvers, runs simulations, and produces post-processed results---compressing a multi-role pipeline into a largely one-click, traceable workflow.

\vspace{0.4em}
\textbf{Underlying stack and tools}

FlowXMaster is implemented as a multi-agent CFD system on Bohrium. Core elements include:
\begin{itemize}[leftmargin=*]
  \item \emph{Sub-agents:} geometry, meshing, solving, post-processing, with basic self-checking and retry.
  \item \emph{Executable services:} modelling, meshing, CFD, and visualisation exposed with unified job and file semantics.
  \item \emph{Representative stack:} GLM-4.5V+\texttt{rembg}, Meshy APIs, OpenFOAM (\texttt{blockMesh}, \texttt{snappyHexMesh}), scripted ParaView.
\end{itemize}

\vspace{0.4em}
\textbf{Efficiency and impact}

FlowXMaster externalises CFD setup into an agent-driven workflow, reducing configuration overhead and enabling minutes-scale iteration for early-stage engineering design and optimisation.

\vspace{0.4em}
\textbf{Representative case}

\textit{Image-to-external flow-field simulation.}
Given a single automotive image, FlowXMaster segments the object, reconstructs a 3D model, runs an OpenFOAM case, and outputs flow-field visualisations and key indicators (e.g., velocity and pressure distributions).

\end{agentcard}

\clearpage
\begin{agentcard}[MatMaster]

\textbf{Identity}

\agentfield{Name}{MatMaster}
\agentfield{Domain}{Materials science and formulation design}
\agentfield{Maturity}{Production}
\agentfield{Tagline}{Closed-loop materials design across reading, computation, and experiment.}

\vspace{0.35em}
\textbf{Problem and workflow}

\textit{Traditional bottleneck.}
Materials formulation and process optimization are typically fragmented across
literature review, computational screening, and experimental validation, leading
to slow trial-and-error cycles and high invalid-experiment rates in large,
multi-parameter design spaces.

\textit{Agentic workflow.}
MatMaster implements an end-to-end \emph{Read--Compute--Experiment} loop. It
constructs structured formulation and process spaces from literature and
historical data, performs large-scale computational or learning-based screening,
proposes candidate designs, and—when laboratory access is available—executes
experimental validation. Experimental outcomes are re-ingested to guide
subsequent iterations, forming a traceable dry--wet optimization cycle.

\vspace{0.35em}
\textbf{Underlying stack and execution model}

MatMaster runs as a production materials agent on Bohrium:
\begin{itemize}[leftmargin=*]
  \item \emph{Knowledge grounding:} large-scale literature and data mining for
        formulation, process, and performance variables.
  \item \emph{Computational screening:} simulation- and model-driven exploration
        of high-dimensional chemical and process spaces.
  \item \emph{Closed-loop execution:} experiment design, laboratory execution,
        and quantitative analysis with feedback to subsequent screening.
\end{itemize}

\vspace{0.35em}
\textbf{Efficiency and impact}

Across multiple materials design scenarios, MatMaster achieves order-of-magnitude
improvements in screening efficiency, reduces invalid experiments by up to
$\sim80\%$, and shortens optimization cycles from months to days, while remaining
physically and experimentally grounded.

\vspace{0.35em}
\textbf{Representative cases}

\textit{Electrolyte and perovskite materials design.}
MatMaster has been applied to electrolyte formulation optimization and
perovskite solar-cell additive and process co-design, combining literature
mining, large-scale virtual screening, and experimental validation to converge
on improved material performance within a small number of agent-driven
iterations.

\end{agentcard}

\clearpage
\begin{agentcard}[ML-Master]

\textbf{Identity}

\agentfield{Name}{ML-Master}
\agentfield{Domain}{Machine learning experimentation}
\agentfield{Maturity}{Prototype}
\agentfield{Tagline}{From task description to executable ML experiments.}

\vspace{0.4em}
\textbf{Problem and workflow}

\textit{Traditional bottleneck.}
End-to-end machine learning experiments involve long chains of data processing,
model design, training, debugging, evaluation, and iteration. Much of this work
is manual, repetitive, and error-prone, with experimental knowledge fragmented
across scripts, logs, and ad hoc notes, making systematic exploration and reuse
difficult.

\textit{Agentic workflow.}
Given a task description and evaluation criteria, ML-Master executes a closed-loop
\emph{Plan--Code--Run--Improve} workflow. A research agent proposes executable
experiment plans grounded in prior knowledge, while a coding agent implements,
runs, debugs, and iteratively refines code under a fixed time budget, producing
traceable and reproducible ML experiments rather than isolated trials.

\vspace{0.4em}
\textbf{Underlying stack and execution model}

ML-Master is implemented as a dual-agent ML system on Bohrium:
\begin{itemize}[leftmargin=*]
  \item \emph{Research agent:} task parsing, method retrieval, proposal generation,
        and maintenance of experiment context.
  \item \emph{Coding agent:} code generation, execution, debugging, and result-driven
        modification within the execution loop.
  \item \emph{Execution substrate:} model training, evaluation, and logging executed
        as governed, replayable jobs on Bohrium’s compute infrastructure.
\end{itemize}

\vspace{0.4em}
\textbf{Efficiency and impact}

ML-Master compresses multi-day or multi-week human-driven ML experimentation into
hours-scale autonomous exploration, enabling hundreds of executable iterations
within a single budgeted run while maintaining stable execution and reproducible
outputs.

\vspace{0.4em}
\textbf{Representative case}

\textit{Kaggle-scale ML engineering tasks.}
On realistic ML benchmarks such as MLE-Bench and Kaggle-style competitions,
ML-Master achieves competitive or state-of-the-art performance within a
12-hour execution budget, demonstrating robust experiment planning, execution,
and refinement without human intervention.

\end{agentcard}

\clearpage
\begin{agentcard}[OPT-Master]

\textbf{Identity}

\agentfield{Name}{OPT-Master}
\agentfield{Domain}{Optimization and Operations Research}
\agentfield{Maturity}{Prototype}
\agentfield{Tagline}{From problem description to executable optimization strategies.}

\vspace{0.35em}
\textbf{Problem and workflow}

\textit{Traditional bottleneck.}
Many real-world optimization problems suffer from a dual bottleneck:
natural-language problem descriptions must be manually translated into
mathematical formulations, and NP-hard problems rely on hand-crafted
heuristics that are brittle, domain-specific, and costly to tune.

\textit{Agentic workflow.}
OPT-Master runs a structured \emph{Formulate--Search--Refine} loop.
It interprets problem descriptions, constructs explicit optimization models
(objectives, variables, constraints), and explores solution strategies by
encoding solvers and heuristics as executable programs. Candidate strategies
are evaluated in parallel and iteratively refined based on empirical
performance.

\vspace{0.35em}
\textbf{Underlying stack and execution model}

OPT-Master operates on SciMaster+Bohrium with:
\begin{itemize}[leftmargin=*]
  \item \emph{Code-as-Solver:} optimization problems (objectives, variables,
        constraints) expressed as executable programs, encoding both
        mathematical formulations via solver APIs and algorithmic strategies
        such as metaheuristics and neural combinatorial optimization.
  \item \emph{Code-as-Verifier:} numerical verification through deterministic
        code execution rather than natural language reasoning, validating
        constraint satisfaction and solution correctness.
  \item \emph{DAG-based exploration:} candidate strategies evaluated in parallel
        on Bohrium and iteratively refined via multi-parent derivation,
        enabling cross-trajectory knowledge synthesis.
\end{itemize}

\vspace{0.35em}
\textbf{Efficiency and impact}

OPT-Master automates the design and validation of optimization strategies,
reducing expert-intensive modeling and tuning cycles from weeks to hours for
complex NP-hard tasks.

\vspace{0.35em}
\textbf{Representative case}
On OR benchmarks (IndustryOR, MAMO-Complex, OptMATH), OPT-Master consistently
outperforms GPT-5 and specialized methods by significant margins.
On the NP-hard 26-circle packing problem, OPT-Master automatically discovered
a high-quality heuristic that outperformed AlphaEvolve, ShinkaEvolve, and
other agent-based baselines within hours, and generalized to larger instances
($n=32$) without manual redesign.

\end{agentcard}

\clearpage
\begin{agentcard}[PaSaMaster]

\textbf{Identity}

\agentfield{Name}{PaSaMaster}
\agentfield{Domain}{Cross-domain scientific literature retrieval}
\agentfield{Maturity}{Production}
\agentfield{Tagline}{From open-ended questions to traceable, high-recall literature sets.}

\vspace{0.35em}
\textbf{Problem and workflow}

\textit{Traditional bottleneck.}
Scientific literature search remains fragmented and keyword-driven.
Conventional academic search engines favor shallow matching, while
general-purpose LLM assistants can reason deeply but often miss coverage
and lack traceable evidence. As a result, researchers trade recall for depth,
or vice versa.

\textit{Agentic workflow.}
PaSaMaster implements a structured \emph{fast--slow, T-shaped} retrieval loop.
A fast agent rapidly expands candidate papers through citation networks and
semantic similarity, while a slow reasoning agent refines intent, disambiguates
concepts, and re-ranks candidates through deeper analysis. The two operate
in parallel over a shared evidence space, producing both breadth and precision.

\vspace{0.35em}
\textbf{Underlying stack and execution model}

PaSaMaster runs on SciMaster and Bohrium with:
\begin{itemize}[leftmargin=*]
  \item \emph{Large-scale corpora:} arXiv and Science Navigator–scale literature
        repositories.
  \item \emph{Dual-agent coordination:} fast expansion plus slow reasoning
        without blocking.
  \item \emph{Trace-backed ranking:} citation-aware retrieval with explicit
        provenance to original sources.
\end{itemize}

\vspace{0.35em}
\textbf{Efficiency and impact}

PaSaMaster compresses hours of manual, multi-platform literature search
into minutes-scale agent execution, while maintaining high recall and
transparent source attribution across disciplines.

\vspace{0.35em}
\textbf{Representative case}

\textit{Open-ended cross-disciplinary survey seeding.}
Given a natural-language research question, PaSaMaster retrieves and ranks
hundreds of relevant papers from large corpora within minutes, producing a
traceable literature set that directly feeds downstream agents such as
SurveyMaster and domain modeling workflows.

\end{agentcard}

\clearpage
\begin{agentcard}[PDEMaster]

\textbf{Identity}

\agentfield{Name}{PDEMaster}
\agentfield{Domain}{Computational mechanics and applied mathematics}
\agentfield{Maturity}{Prototype}
\agentfield{Tagline}{From natural-language PDE descriptions to validated simulations.}

\vspace{0.35em}
\textbf{Problem and workflow}

\textit{Traditional bottleneck.}
Finite-element PDE simulation requires expertise in physical modelling, weak-form
derivation, meshing, and solver configuration. Translating equations from papers
into executable code is repetitive and error-prone, with non-convergence and
unphysical results often requiring days of manual debugging.

\textit{Agentic workflow.}
PDEMaster executes a \emph{Read--Discretize--Simulate--Validate} loop. Starting from
a natural-language problem description, it retrieves relevant PDE formulations
and numerical patterns, constructs meshes and variational forms, runs simulations
on Bohrium, and evaluates results through physics-based consistency checks. When
violations are detected, the agent autonomously adjusts discretization or solver
parameters and re-runs the simulation.

\vspace{0.35em}
\textbf{Underlying stack and execution model}

PDEMaster operates as a simulation-centric agent on Bohrium:
\begin{itemize}[leftmargin=*]
  \item \emph{Model grounding:} retrieval of weak forms, constitutive relations,
        and solver patterns from literature and documentation.
  \item \emph{Numerical execution:} automated meshing, finite-element assembly,
        and nonlinear solving executed as governed compute jobs.
  \item \emph{Closed-loop validation:} residual analysis, boundary-condition
        checks, and qualitative field inspection driving self-correction.
\end{itemize}

\vspace{0.35em}
\textbf{Efficiency and impact}

PDEMaster reduces time-to-simulation from multi-day expert workflows to
hour-scale execution, shifting researchers’ effort from low-level code debugging
to high-level physical modelling and interpretation.

\vspace{0.35em}
\textbf{Representative case}

\textit{Autonomous phase-field fracture simulation.}
Given a textual description of a single-edge-notched specimen and an AT2 phase-
field model, PDEMaster generated a finite-element simulation, detected
nonphysical crack smearing, adjusted discretization parameters, and converged to
a physically consistent fracture path without human intervention.

\end{agentcard}

\clearpage
\begin{agentcard}[PharmMaster]

\textbf{Identity}

\agentfield{Name}{PharmMaster}
\agentfield{Domain}{Medicinal chemistry and patent analysis}
\agentfield{Maturity}{Production}
\agentfield{Tagline}{A patent-aware AI co-scientist for medicinal chemistry.}

\vspace{0.35em}
\textbf{Problem and workflow}

\textit{Traditional bottleneck.}
Medicinal chemists face severe friction in patent-driven research.
Patent documents are long and heterogeneous; Markush structures are difficult
to enumerate; structure--activity relationships (SAR) require expert synthesis;
and freedom-to-operate (FTO) analysis is costly and error-prone.
As a result, patent intelligence is often delayed until late-stage design.

\textit{Agentic workflow.}
PharmMaster runs an integrated \emph{Read--Analyze--Assess} loop.
It parses patent documents into structured chemical and activity representations,
automatically reconstructs Markush scopes, aggregates SAR signals across patents,
and evaluates candidate molecules against patent protection boundaries.
The workflow produces traceable reports grounded in explicit patent evidence.

\vspace{0.35em}
\textbf{Underlying stack and execution model}

PharmMaster operates on SciMaster+Bohrium with:
\begin{itemize}[leftmargin=*]
  \item \emph{Patent parsing:} structured extraction of text, molecules,
        activities, and Markush descriptions.
  \item \emph{Chemical reasoning:} scaffold decomposition, R-group analysis,
        and cross-patent SAR synthesis.
  \item \emph{Risk assessment:} candidate--Markush matching and FTO evaluation
        with explicit provenance.
\end{itemize}

\vspace{0.35em}
\textbf{Efficiency and impact}

PharmMaster compresses patent landscaping from $\sim$10 days to under one day.
Single-molecule FTO assessment is reduced from multi-day expert analysis to
minutes-scale execution, while avoiding false-positive risk flags in practice.

\vspace{0.35em}
\textbf{Representative case}

\textit{BTK target patent landscape and FTO analysis.}
Given a BTK target query and a candidate molecular structure, PharmMaster
screened over one million patents, reconstructed Markush protection scopes,
built a cross-patent SAR matrix, and assessed infringement risk.
A workflow that traditionally required $\sim$10 days of expert effort was
completed within one day, producing a structured, traceable BTK patent report.

\end{agentcard}

\clearpage
\begin{agentcard}[PhysMaster]

\textbf{Identity}

\agentfield{Name}{PhysMaster}
\agentfield{Domain}{Theoretical \& Computational Physics}
\agentfield{Maturity}{Near-production}
\agentfield{Tagline}{Theory-grounded, computation-heavy physics workflows at scale.}

\vspace{0.35em}
\textbf{Problem and workflow}

\textit{Traditional bottleneck.}
Many physics tasks require long chains from assumptions and literature to
large-scale computation, repeated fitting and parameter scans, and
physics-consistency checks. These workflows are fragile and time-consuming,
often requiring weeks to months of expert effort.

\textit{Agentic workflow.}
PhysMaster runs a structured \emph{Clarify--Compute--Check} loop: it extracts
assumptions and target observables, grounds the task in relevant literature,
executes large-scale numerical or symbolic computations on Bohrium, and iterates
through self-checking and refinement until results satisfy explicit physical
constraints and reproducibility requirements.

\vspace{0.35em}
\textbf{Underlying stack and execution model}

PhysMaster operates as a multi-agent workflow on SciMaster+Bohrium:
\begin{itemize}[leftmargin=*]
  \item \emph{Task structuring:} clarify assumptions, constraints, and observables.
  \item \emph{Knowledge grounding:} retrieve and consolidate relevant prior work.
  \item \emph{Computation at scale:} fitting, scans, and analysis executed as
        traceable jobs on Bohrium.
\end{itemize}

\vspace{0.35em}
\textbf{Efficiency and impact}

PhysMaster compresses month-scale physics analysis pipelines to day-scale
execution in computation-dominated tasks, while preserving traceability and
physics-based verification.

\vspace{0.35em}
\textbf{Representative case}

\textit{Lattice QCD extraction of the Collins--Soper kernel.}
PhysMaster automated large-scale correlator fitting and parameter scans,
performed required transformations and consistency checks, and extracted the
Collins--Soper kernel. A workflow that previously required $\sim$1 month of
manual effort was completed in under one day with results consistent with expert
analyses.

\end{agentcard}

\clearpage
\begin{agentcard}[SpecMaster]

\textbf{Identity}

\agentfield{Name}{SpecMaster}
\agentfield{Domain}{Spectroscopy and molecular structure elucidation}
\agentfield{Maturity}{Prototype}
\agentfield{Tagline}{Automated structure elucidation from multi-spectral evidence.}

\vspace{0.35em}
\textbf{Problem and workflow}

\textit{Traditional bottleneck.}
Molecular structure elucidation from NMR, MS, and related spectra is highly
expert-dependent. Signals are fragmented across spectral modalities, hypothesis
testing is iterative and manual, and integrating simulation, retrieval, and
chemical rules often takes days even for experienced spectroscopists.

\textit{Agentic workflow.}
SpecMaster runs a \emph{Hypothesize--Simulate--Repair--Validate} loop. Starting
from experimental spectra and optional molecular formula constraints, it
extracts spectral features, proposes candidate structures, performs forward
spectral simulation, retrieves reference structures, and enforces chemical and
geometric rules. When no initial candidate is consistent with the evidence,
SpecMaster iteratively repairs molecular graphs until multi-spectral
consistency is achieved.

\vspace{0.35em}
\textbf{Underlying stack and execution model}

SpecMaster operates as a multi-agent reasoning system on Bohrium:
\begin{itemize}[leftmargin=*]
  \item \emph{Evidence grounding:} peak extraction, spectral pattern recognition,
        and constraint construction across NMR and MS.
  \item \emph{Forward verification:} simulation of chemical shifts and fragment
        patterns to test structural hypotheses.
  \item \emph{Structure repair:} graph-level modification guided by chemical
        rules, geometric plausibility, and spectral consistency.
\end{itemize}

\vspace{0.35em}
\textbf{Efficiency and impact}

SpecMaster compresses day-scale, expert-driven spectral interpretation into
minute-scale automated workflows, enabling high-throughput structure
elucidation with traceable reasoning rather than manual trial-and-error.

\vspace{0.35em}
\textbf{Representative case}

\textit{Structure recovery beyond initial candidates.}
Given experimental $^1$H/$^{13}$C NMR spectra and a molecular formula, SpecMaster
recovered the correct molecular structure even when it was absent from the
initial candidate set, by iteratively repairing incorrect bond topologies until
simulated spectra matched experimental evidence across modalities.

\end{agentcard}

\clearpage
\begin{agentcard}[SurveyMaster]

\textbf{Identity}

\agentfield{Name}{SurveyMaster}
\agentfield{Domain}{Cross-domain literature review}
\agentfield{Maturity}{Production}
\agentfield{Tagline}{From topic specification to executable, citation-grounded surveys.}

\vspace{0.4em}
\textbf{Problem and workflow}

\textit{Traditional bottleneck.}
High-quality scientific surveys require exhaustive literature retrieval, careful filtering, conceptual organisation, and rigorous citation management. These steps are time-consuming, error-prone, and highly dependent on individual expertise, often stretching over weeks or months of manual effort.

\textit{Agentic workflow.}
Given a topic description, SurveyMaster executes a structured \emph{Search--Analyse--Write} workflow: large-scale semantic and citation-aware retrieval, extraction of structured literature cards, clustering and outline refinement, and section-level drafting with explicit citation grounding. The result is a coherent, trace-backed survey pipeline rather than a one-off document.

\vspace{0.4em}
\textbf{Underlying stack and execution model}

SurveyMaster operates as a reading-centric, multi-agent workflow on Bohrium. Core elements include:
\begin{itemize}[leftmargin=*]
  \item \emph{Reading orchestration:} automated query generation and iterative retrieval over large corpora via Science Navigator and PaSaMaster-style reading agents.
  \item \emph{Structured analysis:} extraction of problems, methods, datasets, and conclusions into reusable literature cards, followed by clustering and outline alignment.
  \item \emph{Grounded writing:} section-wise generation with explicit citation linking, producing directly compilable survey drafts.
\end{itemize}

\vspace{0.4em}
\textbf{Efficiency and impact}

SurveyMaster compresses survey production from month-scale expert labour to hours-scale execution, freeing researchers from exhaustive search, annotation, and formatting, and allowing them to focus on critical interpretation and original insight.

\vspace{0.4em}
\textbf{Representative case}

\textit{Automated survey on deep potentials in materials science.}
Starting from a topic description, SurveyMaster retrieved and analysed $\sim10^3$ papers on deep potentials and related workflows, constructed a structured outline, and generated a 70+ page, citation-grounded survey covering theory, methodologies, applications, and open challenges.

\end{agentcard}